\documentclass{article}
\usepackage[utf8]{inputenc}
\usepackage[T1]{fontenc}
\usepackage{geometry}
\usepackage{translator}
\usepackage[catalan, spanish, british]{babel}
\usepackage{johd}

\usepackage{setspace}
\usepackage[colorinlistoftodos]{todonotes}
\usepackage{hyperref}
\usepackage{soul}
\usepackage{underscore}
\usepackage{array, multirow}
\usepackage{tabularx, tabulary, longtable}
\usepackage{tipa}
\usepackage{multicol}
\usepackage{makecell}
\usepackage{booktabs}
\usepackage{enumitem}
\usepackage{ragged2e}
\usepackage{xspace}
\usepackage{etoolbox}

\newcolumntype{Y}{>{\RaggedRight\arraybackslash}X}

\let\oldbibliography\thebibliography
\renewcommand{\thebibliography}[1]{%
  \oldbibliography{#1}%
  \setlength{\itemsep}{8pt}%
  \setlength{\baselineskip}{11.5pt}%
}

\newcommand{\mapu}[1]{\emph{#1}}
\newcommand{\mapuex}[1]{\footnotesize\emph{#1}}
\newcommand{\gloss}[1]{\footnotesize\textsc{\lowercase{#1}}}
\newcommand{\example}[1]{\textbf{E.#1}\xspace}

\usepackage{titlesec}
\titleformat{\section}
  {\normalfont\Large\bfseries}
  {\thesection}{1em}{}
\titleformat{\subsection}
  {\normalfont\large\bfseries}
  {\thesubsection}{1em}{}
\titleformat{\subsubsection}
  {\normalfont\normalsize\bfseries}
  {\thesubsubsection}{1em}{}

\hypersetup{
    colorlinks=true,
    linkcolor=blue,
    citecolor=violet,
    urlcolor=blue,
    pdftitle={Valency Classification of Mapudüngun Verbal Roots},
    pdfauthor={Andrés Chandía}
}

\begin{document}

\begin{center}
  {\Huge\bfseries Valency Classification of \mapu{Mapudüngun} Verbal Roots\par}
  \vspace{1em}
  {\Large\textit{Established by the language's own morphotactics}\par}
  \vspace{2cm}
  {\large
  Andrés Chandía\\[2pt]
  \small\href{mailto:andres@chandia.net}{\texttt{andres@chandia.net}}\\[8pt]
  Department of Catalan Philology and General Linguistics\\ 
  University of Barcelona\\[4pt]
  PhD programme: Cognitive Science and Language\\[12pt]
  {\small Supervised by}\\
  {\small Dr. Elisabet Comelles \href{mailto:elicomelles@ub.edu}{\texttt{(elicomelles@ub.edu)}} and\\
  {\small Dr. Irene Castellón \href{mailto:icastellon@ub.edu}{\texttt{(icastellon@ub.edu)}}}}
  }
  \vfill
\end{center}
\thispagestyle{empty}

\newpage

\begin{abstract}
In the previous work, a lexical (re)categorisation —or confirmation of the given category— of roots identified as verbal was undertaken to determine their original category accurately. Building on this, the present paper offers an account of the valency classification of those \mapu{Mapudüngun} roots confirmed to be verbal, using the language's own morphotactics; specifically, by examining the permissible and restricted combinations of various suffixes with roots or verbal stems in the \mapu{Mapuche} verb form.

As with all work conducted thus far, the results presented here aim to improve the morphological analyser (\mapu{Düngupeyüm}) with all verified findings incorporated into the system. From a theoretical perspective, we also hope to contribute to the recognition and understanding of issues related to the valency of \mapu{Mapuche} verb forms.
\end{abstract}

\noindent\textbf{Keywords:} Mapuche; Mapudüngun; FST; finite state transducer; morphology; morphosyntax; affixes; suffixes; morphemes; verbs; valency; verbal roots; parts of speech

\vspace{1cm}
\selectlanguage{spanish}
\begin{abstract}
En el trabajo inmediatamente anterior, se llevó a cabo una (re)categorización léxica —o confirmación de la categoría asignada— de raíces identificadas como verbales, para determinar con precisión su categoría original. Sobre esta base, el presente artículo ofrece una descripción de la clasificación valencial de aquellas raíces del \mapu{Mapudüngun} confirmadas como verbales, mediante la propia morfotáctica de la lengua; específicamente, examinando las combinaciones permitidas y restringidas de diversos sufijos con raíces o temas verbales que ocurren en la forma verbal \mapu{Mapuche}.

Como todo el trabajo realizado hasta ahora, los resultados presentados aquí tienen como objetivo mejorar el analizador morfológico (\mapu{Düngupeyüm}) incorporando todos los temas estudiados al sistema. Desde una perspectiva teórica, también esperamos contribuir al reconocimiento y comprensión de cuestiones relacionadas con la valencia de las formas verbales \mapu{Mapuche}.
\end{abstract}

\noindent\textbf{Palabras clave:} Mapuche; Mapudüngun; FST; transductor de estados finitos; morfología; morfosintaxis; afijos; sufijos; morfemas; verbos; valencia; raíces verbales; categorías gramaticales

\vspace{1cm}
\selectlanguage{catalan}
\begin{abstract}
En el treball immediatament anterior, es va dur a terme una (re)categorització lèxica —o confirmació de la categoria assignada— de les arrels identificades com a verbals, per a determinar amb precisió la seva categoria original. A partir d'allò, el present article ofereix una descripció de la classificació de la valència d'aquelles arrels del \mapu{Mapudüngun} confirmades com a verbals, mitjançant la morfotàctica pròpia de la llengua; específicament, examinant les combinacions permeses i restringides de diversos sufixos amb arrels o temes verbals que apareixen en la forma verbal \mapu{Maputxe}.

Com tot el treball realitzat fins ara, els resultats presentats aquí tenen com a objectiu millorar l'analitzador morfològic (\mapu{Düngupeyüm}) incorporant tots els temes estudiats en el sistema. Des d'una perspectiva teòrica, també esperem contribuir al reconeixement i comprensió de qüestions relacionades amb la valència de les formes verbals \mapu{Maputxe}.
\end{abstract}

\noindent\textbf{Paraules clau:} Maputxe; Mapudüngun; FST; transductor d'estats finits; morfologia; morfosintaxi; afixos; sufixos; morfemes; verbs; valència; arrels verbals; categories gramaticals

\newpage
\selectlanguage{british}
\tableofcontents

\subsection*{Note on the spelling of \mapu{Mapuche} names}

In this article, we respect the spelling of \mapu{Mapuche} names and surnames according to the \textit{alfabeto mapuche unificado} (AMU), as an act of recognition and respect towards the \mapu{Mapuche} people and their language. Therefore, we use the forms \mapu{Longkon} and \mapu{Koña} instead of the standard bibliographical Castilianisations (Loncón, Coña). In the bibliographic references, however, the standardised forms are added in parentheses to facilitate the location of the sources.
\vspace{2cm}

\section{\label{sec.01} Introduction}

In this article, we present the methodologies employed for the reclassification of the valency of \mapu{Mapuche} verb roots (see \S\ref{sec.02}), following a prior reclassification of lexical roots \citep{chandia2025}, and the outcomes of this process. This article further delineates the intricacies involved in reclassifying the valency of roots categorised as verbal (see \S\ref{sec.02.2}) in accordance with the classification established by Ineke \citeauthor{smeets2008} in her \citeyear{smeets2008} publication, \textit{``A Grammar of Mapuche''}. Additionally, it encompasses roots that were identified as verbal in the preceding lexical analysis. The reader is invited to peruse the initial article \citep{chandia2025}, which is concise and establishes the foundation for the work that led to the creation of both documents.

As with all the research related to the \mapu{Düngupeyüm}\footnote{Available on \url{https://www.chandia.net/dungupeyum/}.\vspace{0.2cm}} project and carried out to improve our computational morphological analysis system (\citealp{chandia2012}, \citeyear{chandia2021}, \citeyear{chandia2022}), this work is based on \cite{smeets2008}'s grammatical description. The system's initial implementation was totally based on her work. The accurate determination of verb root valency is crucial to this computational system, while valency influences the composition of verb forms, which also include suffixation and morphological changes. This factor is essential for correctly identifying the suffixes present in a verb form.

During the tool's development, several observations by \cite{smeets2008} were verified and modified. Her description of \mapu{Mapudüngun} grammar, based on data from ten informants, does not fully capture the language's diversity or all its phenomena. In any case, \cite{smeets2008}'s \textit{``A Grammar of Mapuche''} significantly contributes to computational morphological analysis by identifying a fixed position for about one hundred suffixes in \mapu{Mapuche} verb forms and providing their morphotactics.

Following \cite{smeets2008}'s description, suffixes occupy relatively fixed positions, with verbal affixes arranged in thirty-six numbered slots, from right to left. Some slots contain mutually exclusive suffixes due to grammatical or semantic reasons. In \mapu{Mapudüngun}, a minimal verb form starts with a root, some optional derivational suffixes attached and one inflectional suffix (sometimes two or three). A mood marker is required in slot four for finite forms, as well as a person marker in slot three. While up to seven or eight suffixes can follow the root (see E.1 and Figure 1 in \citealp{chandia2025}), verbs typically have between four and six suffixes in spontaneous speech.

\cite{smeets2008}'s classification of verb roots by valency, based on recorded usage, is now considered potentially inaccurate. As is common knowledge in the field, a considerable number of verbs that traditionally fall into the transitive category are observed in intransitive contexts, and conversely, a significant proportion of verbs that traditionally fall into the intransitive category can also be found in transitive contexts (\citealp{dedios2013}: 425). In \mapu{Mapudüngun}, as in other agglutinative languages, suffixes can transform verb valency, making them either transitive or intransitive regardless of their original valency. Semantic classification\footnote{\cite{zuniga2024} offers a semantic interpretation not for classifying verb roots, but for explaining the selection of applicative suffixes, including \mapu{-tu}. In this work, the suffix \mapu{-tu} is discussed from a morphological perspective in \S\ref{sec.03.7}.\vspace{0.2cm}} may not yield a single valency for each verb, so the most frequent valency is used. Syntactic classification describes how verbal forms behave in sentences, which can vary depending on the context\footnote{\label{gollzuni}\citeauthor{golluscio2010} (\citeyear{golluscio2007}, \citeyear{golluscio2010}), \citeauthor{zuniga2006} (\citeyear{zuniga2006}, \citeyear{zuniga2009b}, \citeyear{zuniga2011}, \citeyear{zuniga2024}) provide an in-depth study of the valency of \mapu{Mapuche} verbs from a syntactic perspective.\vspace{0.2cm}}.

In \mapu{Mapudüngun}, behaviour is marked by morphosyntactic indices due to its polysynthetic nature\footnote{``... SYNTHESIS refers to how many meaningful morphemes can combine to form a word in a language. If there are few ... it is referred to as analysis. If there are more ... it is synthesis. If there are many ... it is polysynthesis'' (\citealp{zuniga2006}: 199).\vspace{0.2cm}}. Being an agglutinative\footnote{'Agglutination' is a morphological pattern in which morphemes appear inside a word without changing form in different contexts, and distinct semantic elements remain clearly distinguishable (\citealp{zuniga2006}: 200).\vspace{0.2cm}} language, \mapu{Mapudüngun} uses suffix clusters to represent what other languages might express with phrases or sentences. Thus, verb valency is influenced by suffixes, sentence elements, and semantic value of all of them\footnote{\cite{tubino2017} provides an excellent explanation related to the topics covered in this article, including the associated terms. In the introduction, she highlights the interweaving of morphology and syntax in agglutinative languages. In section two, she defines and describes verbal valency and introduces valency change processes in Spanish and English. In section three, the author explains these processes in \textit{Yaqui} —a language that, like \mapu{Mapudüngun}, is also agglutinative— and the affixes involved in these processes.\vspace{0.2cm}}.

This work employs morphology, with a particular focus on the language's inherent morphotactics, to ascertain verbal valency. Although semantics and syntax have been employed on occasion to address ambiguities (see \S\ref{sec.02.2}), these techniques were applied only superficially.

\subsection{\label{sec.01.1} The Study of Verbal Valency}

In recent decades, significant progress has been made in the field of verb valency classes, with numerous distinguished linguists contributing to this area of research. \textit{Valency Classes in the World's Languages} \citep{malchukov-comrie2015} compiles an investigation into the argument-structure properties of verbs of different valency classes from a typological perspective. The objective is to ascertain general patterns in verb syntax through an empirical examination of a cross-linguistically comparable list of verbs grounded semantically, rather than syntactically, and limited to verb meanings which are likely to have equivalents across languages. This investigation covers a set of 70 core verb meanings across 30 languages, incorporating a \mapu{Mapudüngun} study conducted by \cite{zuniga2011}.

Following \cite{creissels2015}, valency is understood here as the number of obligatory arguments a verb requires, distinguished from the number of syntactic positions it may optionally license. \mapu{Mapudüngun} does not exhibit a clear transitivising or intransitivising profile cross-linguistically: morphologically, causatives transitivise, but syntactically there are labile alternations. This mixed pattern is addressed throughout the analysis.

As previously mentioned, this research is undertaken through the semantics of the verbs, and for many of the languages studied, their morphology reflects either the valency the verb turns into, or the valency needed by a specific affix to participate. Building on the findings of this research on \mapu{Mapudüngun}, we have undertaken a process akin to what is termed ``reverse engineering'' in computational fields. By analysing the morphology inherent in the forms, we have sought to determine whether a given verb root is transitive, intransitive, or labile, with the aim of identifying its most fundamental —and ideally unique— valency.

\subsection{\label{sec.01.2} Mapudüngun: Morphosyntactic and Sociolinguistic Profile}

\mapu{Mapudüngun}\footnote{The spelling \mapu{Mapudüngun} follows the \textit{alfabeto mapuche unificado} (AMU), where \textit{ü} represents the central vowel /\textschwa/, which is extremely short in \mapu{Mapudüngun}, sometimes almost imperceptible. The pronunciation \textit{düngu} is attested historically (Valdivia, Febrés, Augusta) and in contemporary sources \citep{navarro1993}.\vspace{0.2cm}} is a polysynthetic and predominantly suffixing language spoken by approximately 200,000 people in Chile and Argentina \citep{zuniga2006}. It is classified as endangered by UNESCO. The language exhibits split ergativity based on tense and person, free word order with a tendency toward SVO/AOV, and a rich system of valency-changing operations. Verbal morphology is templatic: roots are followed by derivational suffixes, then inflectional markers for mood, person, and number.

\section{\label{sec.02} Methodology}

As noted in \S\ref{sec.01}, accurately identifying the valency of verb roots is crucial for the computational system developed. This is essential for morphological analysis, as many rules governing root and suffix combinations depend on this classification or valency changes during word formation. This ensures an unambiguous analysis or one with minimal ambiguity (see footnote 14 in \citealp{chandia2025}).

In the future, it may be possible to dispense with this classification, which some researchers (\citealp{dedios2013}: 425) consider outdated. To do so, the system would need to be reconfigured, after classifying \mapu{Mapudüngun} verb roots from alternative perspectives. This process would enable the development of morphotactics and morphosyntactic combination rules based on different constraints. For now, the valency classification of verbs will continue to be integrated within the developed computational system.

\subsection{\label{sec.02.1} Lexical Category of the Roots}

The verb roots listed by \cite{smeets2008} in her dictionary were taken to verify their valency classification and to modify it if necessary. First, it was ensured that these were indeed verb roots and not verbalised forms of other lexical categories. This task is explained in \cite{chandia2025} with the results presented in Table 3 of that work.

\subsection{\label{sec.02.2} Valency of Verbal Roots}

In a subsequent phase of the study detailed in this article, the roots belonging to the verb class are classified based on their valency. However, it is possible that some roots may have been misidentified and are, in fact, part of a different lexical category. This may occur because in \mapu{Mapudüngun}, roots belonging to categories different than the verbal one are frequently verbalised, which can lead to uncertainty regarding the initial category of the root.

The review will begin examining the roots classified as intransitive to verify whether they truly express this valency or the transitive one. This classification is not straightforward and is even complicated in \mapu{Mapudüngun}. As \citeauthor{deaugusta1903} (\citeyear{deaugusta1903}: 294) noted, ``Some neutral verbs\footnote{Neutral verbs are those commonly referred to today as `intransitive'.\vspace{0.2cm}} are also used as transitive of the 1\textsuperscript{st} class''. On the one hand, the Capuchin was referring to the final form of the verb, since some suffixes add a semantic-grammatical nuance that makes transitive verb forms containing intransitive roots (there are also labile roots, which will be addressed in \S\ref{sec.03.4.1}).

Below are some examples of roots that are evidently intransitive\footnote{In \mapu{Mapudüngun}, intransitive verbs are often translated using the pronominal form in Spanish when the verb is not intransitive in its simple form in Spanish. For example, \mapu{wüyü-} `marearse -- get dizzy', \mapu{allfü-} `dañarse -- get damaged', \mapu{ñuwi-} `perderse -- get lost', etc.\vspace{0.2cm}} but can form transitive verbal predicates. \hyperref[e1a]{E.1a} and \hyperref[e1c]{E.1c} are intransitive, while \hyperref[e1b]{E.1b} and \hyperref[e1d]{E.1d} are transitivised by the causative morphemes \mapu{-(ü)l} and \mapu{-(ü)m} respectively.

\paragraph{\example{1} \label{e1}}~
\vspace{-10pt}
\begin{enumerate}[label=\alph*.]
\item \label{e1a} \mapuex{küpa \hspace{18pt} -n} \hfill (\citealp{deaugusta1903}: 295)\\
      \gloss{IV.come +IND1SG}\\
      `I came'
\item \label{e1b} \mapuex{küpa \hspace{18pt} \textbf{-l} \hspace{10pt} -ün}\\
      \gloss{IV.come \textbf{+CA} +IND1SG}\\
      `I brought'; \textit{literal:} `I made come'
\item \label{e1c} \mapuex{püra \hspace{22pt} -n} \hfill (\citealp{deaugusta1903}: 296)\\
      \gloss{IV.go-up +IND1SG}\\
      `I went up'
\item \label{e1d} \mapuex{püra \hspace{20pt} \textbf{-m} \hspace{8pt} -ün}\\
      \gloss{IV.go-up \textbf{+CA} +IND1SG}\\
      `I lifted up'; \textit{literal:} `I made go up'
\end{enumerate}

On the other hand, as \citeauthor{moreno2000} (\citeyear{moreno2000}: 353-370) explains and was introduced in \S\ref{sec.01}, verbs can be classified from three perspectives: semantic, syntactic, and morphological. The latter approach, which will be discussed in more detail in \S\ref{sec.03}, involves identifying those suffixes that combine exclusively with verb roots of a particular valency. This pattern has been observed in \mapu{Mapudüngun}.

Syntactic analysis allows us to observe the valency reflected by a verb when it interacts with other elements in a sentence. This means that depending on how the sentence is constructed and the lexical items it includes, a verb will behave as either transitive or intransitive (see footnote \ref{gollzuni}). In terms of semantics, we can apply this classification criterion to a verb based on the phrase it appears in, in which case it should align with the syntactic projection. However, it's also possible to analyse the verb form in isolation. As explained in \S\ref{sec.01}, the valency value obtained can be unique to that form, the most frequent for that form, or it could be determined that the verb is labile—behaving as either transitive or intransitive with similar frequency. Obviously, this depends on the syntactic realisation of the phrase in which the analysed form appears.

\subsection{\label{sec.02.3} Analytic Procedure}

The reverse engineering approach is implemented in five steps: (a) hypothesise a root's valency based on \cite{smeets2008}' classification; (b) systematic combination of the root with diagnostic suffixes (causatives \mapu{-(ü)m}/\mapu{-(ü)l}, progressive \mapu{-nie}, stative \mapu{-(kü)le}, etc.); (c) verification of attested forms in the primary corpus (\mapu{Koña}'s memoirs) and contrastive verification in \cite{smeets2008}; (d) cross-validation with lexical sources (de Augusta, CORLEXIM); (e) resolution of discrepancies by prioritising intransitive interpretations when \cite{smeets2008} and \mapu{Koña} disagree (following the principle that transitivisation is overtly marked, while intransitivity may be unmarked).

\subsection{\label{sec.02.4} Corpus and Sources}

The study draws on a hierarchically organised corpus. The primary corpus consists of \mapu{Koña}'s memoirs (\citealp{mosbach1930}), comprising approximately 450 pages of naturalistic narrative text. \cite{smeets2008}, a descriptive grammar based on elicitation with ten speakers, serves as the contrastive source. Lexical verification sources include de Augusta (1903) and the CORLEXIM online database. Secondary sources include \cite{zuniga2006} and \cite{salas1992a}. When \cite{smeets2008} and \mapu{Koña} disagree on a root's valency, the analysis prioritises \mapu{Koña}'s usage as naturalistic discourse, while noting discrepancies.

\section[Transitivity and Intransitivity in \mapu{Mapudüngun}: Morphological Criteria]{\label{sec.03} Transitivity and Intransitivity in \mapu{Mapudüngun}:\\ \large -- Morphological Criteria --}

In the course of the valency classification of \mapu{Mapuche} roots, reference will be made to various specific aspects of the interaction between affixes and roots that must be given due consideration. Such considerations will primarily involve discussion of the postulates of \cite{smeets2008}, and from these, the development of ideas or explanations regarding the observed behaviour involving the scrutinised elements. The aim of this task is to identify morphological criteria that allow for the recognition of verbal root valency.
   
\subsection{\label{sec.03.1} Suffixes \mapu{-nie}, \mapu{-künu} and \mapu{-(kü)le}}

The suffix \mapu{-nie} adds a sense of durability maintained by the agent of the action. For example, with this suffix, `separate' becomes `keep separated'; the suffix is labelled as `Progressive Persistent', \textsc{PRPS}.

The suffix \mapu{-künu} denotes an activity that has a sense of durability where the agent neither maintains nor interrupts it. For example, with this suffix, `separate' becomes `leave/keep separated' (as it was)\footnote{\label{kununote}We propose that \mapu{-künu} is not a suffix when immediately follows the root, but rather forms a compound verbal theme. In this case, it is the verbal root \mapu{künu-} `leave' that, for example, forms verbs such as \mapu{weyel-kunu} `let swim', \mapu{anü-kunu} `let sit', \mapu{lüg-künu} `let whiten', `whiten', \mapu{treka-künu} `let walk' (\mapu{treka-künu-w-a-n} `I will go for a walk'; \textit{literal:} `I will let myself walk'), etc. The verb \mapu{el-}, also meaning `leave', functions similarly, except that the main verb would be expressed as a participle in translation, for example, \mapu{anü-el-} `leave seated'.\vspace{0.2cm}}; the suffix is labelled as `Perfect Persistent'.

The morpheme \mapu{-(kü)le}\footnote{When suffixes are presented with part of them in parentheses, it indicates that there are two allomorphs of the morpheme, in this case \mapu{-küle} and \mapu{-le}; for other suffixes, such as \mapu{-(ü)m}, the allomorphs would be \mapu{-üm} and \mapu{-m}. The phonological contexts for the different variants generally involve the presence of a preceding vowel or consonant.\vspace{0.2cm}} is labelled `Stative'. Depending on the type of verb, this suffix either adds the notion of a state or an acquired property, or that of an event that persists for at least some time. Some verbs can be interpreted in either way depending on the context in which they occur. For example, \mapu{püra-le-y} can mean both `it is up' and `it is going up'; \mapu{pire-le-y} can mean `it is snowy' or `it is snowing'; \mapu{arof-küle-y} can mean `he is sweaty' or `he is sweating' (examples from \citealp{smeets2008}: 282).

\begin{quote} \label{qt01smeets}
Verbs that contain \mapu{-nie} or \mapu{-künu} are transitive; the action is directed towards the direct object. Verbs that contain \mapu{-(kü)le} are intransitive and do not take suffixes in slot 6. Therefore, verbs that logically require a patient do not take \mapu{-(kü)le} (\citealp{smeets2008}: 169).
\end{quote}

... ``Verbs that logically require a patient''... are transitive verbs. The author notes that \mapu{-(kü)le} does not appear in transitive verbs. This would allow us to recognise as intransitive any verbal root that occurs with the stative suffix \mapu{-(kü)le}, which is debatable, as in her own work, this morpheme is found combined with transitive roots and occurring in transitive verbal forms. See the following examples:

\paragraph{\example{2} \label{e2}}~
\vspace{-10pt}
\begin{enumerate}[label=\alph*.]
\item \label{e2a} \mapuex{elu \hspace{22pt} -w \hspace{5pt} \textbf{-küle} \hspace{1pt} -fu \hspace{6pt} -n \hspace{26pt} ñi \hspace{16pt} tripa \hspace{14pt} -ya \hspace{8pt} -l} \hfill (\citealp{smeets2008}: 292; 21)\\
      \gloss{TV.give +REF \textbf{+ST} \hspace{1pt} +RI +IND1SG -SP.my -IV.leave +FUT +OVN}\\
      `I was ready to leave'; \textit{literal:} `I was about to give me my exit'\footnote{The system returns an alternative analysis of the verbal form that does not contradict its meaning and is also formed with a transitive root: \gloss{TV.el\_set\_put+REF.w+ST.le+RI.fu+IND1SG.n}. With this analysis, the literal translation would be ``[I] was about to \textbf{let me} my exit''.\vspace{0.2cm}}
\item \label{e2b} \mapuex{fey \hspace{28pt} -pi \hspace{12pt} \textbf{-le} \hspace{4pt} -y \hspace{9pt} -ø \hspace{3pt} ta \hspace{2pt} -ti} \hfill (\citealp{smeets2008}: 92; 48)\\
      \gloss{DP.that -TV.say \textbf{+ST} +IND +3 -DP -DP}\\
      `he said so'; \textit{literal:} `That was he saying, yes\footnote{Depending on the context, the form \mapu{tati} can be interpreted in various ways; the phrase containing it does not provide additional context in the source, so its interpretation is somewhat open in this case.\vspace{0.2cm}}'
\end{enumerate}

In \hyperref[e2a]{E.2a}, the verb \mapu{elu-} `give' is transitive, and its use in the phrase is also reflexive (a property provided by the suffix \mapu{-(u)w}). In \hyperref[e2b]{E.2b}, there is pronominal affixation\footnote{Pronominal affixation refers to the marking of object arguments directly on the verb via person agreement suffixes such as \mapu{-fi} and \mapu{-e} (\S\ref{sec.03.5}).\vspace{0.2cm}}; from a syntactic point of view, this could be equivalent to nominal incorporation, and about this, \cite{salas2006} writes:

\begin{quote} \label{qt02salas}
A highly productive construction of verbal themes is the so-called ``direct object incorporation'' or ``object incorporation'': the verbal theme (simple or complex) combines with the noun that serves as its object, and verbal inflection is directly added to the resulting unit (\citealp{salas2006}: 181).
\end{quote}

The object (direct object) identified in this statement confirms the transitivity of the verb in \hyperref[e2b]{E.2b}. Additionally, in this example, we find the transitive verb \mapu{pi-} `say', which provides another argument that contradicts \cite{smeets2008}'s earlier statement. She also mentions that ``verbs containing \mapu{-nie} or \mapu{-künu} are transitive''; this should be interpreted to mean that verbal forms containing these suffixes are transitive, not necessarily the roots themselves, as these could be intransitive roots made transitive. For instance, in \hyperref[e3]{E.3}, the intransitive root \mapu{ngül-} `get together, meet' is part of a verbal form made transitive by the causative (\textsc{CA}, from now on) suffix \mapu{-(ü)m} (see \S\ref{sec.03.3}).

\paragraph{\example{3} \label{e3}}~
\vspace{-10pt}
\begin{enumerate}[label=\alph*.]
\item[] \mapuex{ngül \hspace{50pt} \textbf{-üm \hspace{2pt} -nie} \hspace{8pt} -y \hspace{8pt} -ø \hspace{2pt} plata} \hfill (\citealp{smeets2008}: 111; 10)\\
\gloss{IV.get-together \textbf{+CA +PRPS} +IND +3 -NN}\\
`he saves money'; \textit{literal:} `he makes money gather'
\end{enumerate}
Ultimately, it cannot be determined if a verbal root is transitive when the forms it participates in also include the suffixes \mapu{-nie} or \mapu{-künu}. In \S\ref{sec.04.1}, a detailed examination of these morphemes is provided, along with additional aspects that support this conclusion.

\subsection{\label{sec.03.2} Adjectiviser \mapu{-fal}}

The semantic contribution of the suffix \mapu{-fal} makes the form it attaches to be interpreted as something feasible to do. In \mapu{Düngupeyüm} it is recognised as an ``applicable action adjectiviser'', and labelled `\textsc{ADJDO}' (\textsc{ADJ} for `adjective' and \textsc{DO} for `doable', meaning `feasible').

\citeauthor{smeets2008} (\citeyear{smeets2008}: 312) lists this suffix among the productive derivational nominalizers and explains that forms containing this nominalizer function as adjectives (for this reason it has been reclassified as an adjectiviser in \mapu{Düngupeyüm}). For example, if the verb \mapu{i-} `eat' appears with \mapu{-fal}, it refers to something that is `edible'. Furthermore, it is noted that this suffix only combines with transitive forms; this could be a good way to identify verbs of such valency. Unfortunately, the occurrence of lexical items formed with the adjectiviser \mapu{-fal} is very rare; in the memories of the \mapu{longko Paskwal Koña} \citep{mosbach1930}, there is only one example, \hyperref[e4g]{E.4g}.

An alternative analysis is proposed for the suffix \mapu{-fal} which does not contradict the interpretation of the form containing it. In other words, \mapu{ifal} continues to mean `edible' with the following new analysis—where \hyperref[e4a]{E.4a}, \hyperref[e4c]{E.4c}, \hyperref[e4e]{E.4e} are \cite{smeets2008}'s, and \hyperref[e4b]{E.4b}, \hyperref[e4d]{E.4d}, \hyperref[e4f]{E.4f}, \hyperref[e4g]{E.4g} are the alternative analyses proposed:

\paragraph{\example{4} \label{e4}}~
\vspace{-10pt}
\begin{enumerate}[label=\alph*.]
\item \label{e4a} \mapuex{i \hspace{15pt} \textbf{-fal}} \hfill (\citealp{smeets2008}: 312; 10)\\
      \gloss{eat \textbf{+NOM}} (\textsc{ADJDO} in \mapu{Düngupeyüm})\\
      `edible'
\item \label{e4b} \mapuex{i \hspace{26pt} \textbf{-fa \hspace{18pt} -l} \hspace{12pt} -ø}\\
      \gloss{TV.eat \textbf{+DP.this +CA} +NOM}\\
      `edible'
\item \label{e4c} \mapuex{allkü \hspace{6pt} \textbf{-fal}} \hfill (\citealp{smeets2008}: 312; 11)\\
      \gloss{hear \textbf{+NOM}} (\textsc{ADJDO} in \mapu{Düngupeyüm})\\
      `audible'
\item \label{e4d} \mapuex{allkü \hspace{14pt} \textbf{-fa \hspace{18pt} -l} \hspace{12pt} -ø}\\
      \gloss{IV.hear \textbf{+DP.this +CA} +NOM}\\
      `audible'
\item \label{e4e} \mapuex{pepi \hspace{18pt} -l \hspace{10pt} \textbf{-fal}} \hfill (\citealp{smeets2008}: 312; 12)\\
      \gloss{be-able +CA \textbf{+NOM}} (\textsc{ADJDO} in \mapu{Düngupeyüm})\\
      `feasible, practicable'
\item \label{e4f} \mapuex{pepi \hspace{32pt} -l \hspace{10pt} \textbf{-fa \hspace{18pt} -l} \hspace{10pt} -ø}\\
      \gloss{TV.be-able +CA \textbf{+DP.this +CA} +NOM}\\
      `feasible, possible'
\item \label{e4g} \mapuex{üde \hspace{20pt} \textbf{-fa \hspace{18pt} -l} \hspace{12pt} -ø} \hfill (\citealp{mosbach1930}: 271)\\
      \gloss{TV.hate \textbf{+DP.this +CA} +NOM}\\
      `abominable'
\end{enumerate}

\citeauthor{smeets2008} (\citeyear{smeets2008}: 314) includes the vacuous morpheme \mapu{-ø} in the list of unproductive nominalizers. If the analysis introduced here is correct, this suffix would increase its productivity. According to the alternative analysis, forms ending in \mapu{-fal} are composed of a verbal theme formed by the main verb, which provides the semantic information, and the demonstrative pronoun \mapu{-fa} `this, here (among other possible interpretations)'. The compound becomes a transitive or intransitive verbal theme as it retains the valency of the verbal member of the compound. This verbal theme is followed by the \textsc{CA} \mapu{-(ü)l} and, nominalising (or adjectivising\footnote{In many languages, depending on the syntactic position of the word and/or the usage by the speaker, nominal forms can function adjectivally, and adjectival forms can function nominally. This is also the case in \mapu{Mapudüngun}, but not only for forms whose original category is one of the two, but also for nominalised and adjectivised forms.\vspace{0.2cm}}) the verb, the suffix \mapu{-ø}. According to this analysis, the form \mapu{ifal} could be paraphrased as `this that makes (itself) be eaten', and by extension, `edible'.

The analysis extends to forms ending in \mapu{-l}, a suffix that \citeauthor{smeets2008} (\citeyear{smeets2008}: 313) also includes among the unproductive nominalisers. According to our analysis, it corresponds to the \textsc{CA} rather than to a nominaliser, as shown in the following examples where the first meaning in English is \cite{smeets2008}'s:

\paragraph{\example{5} \label{e5}}~
\vspace{-10pt}
\begin{enumerate}[label=\alph*.]
\item \label{e5a} \mapuex{apo \hspace{42pt} \textbf{-l} \hspace{10pt} -ø}\\
      \gloss{IV.get-filled \textbf{+CA} +NOM}\\
      `filling' -- `aquello lleno, relleno, rellenado' -- `that which is full, stuffed'
\item \label{e5b} \mapuex{ina \hspace{24pt} \textbf{-l} \hspace{8pt} -ø}\\
      \gloss{AV.near \textbf{+CA} +NOM}\\
      `the next', `bank (of river)' -- `el siguiente', la orilla' -- `the next, the following, the shore'
\item \label{e5c} \mapuex{wüne \hspace{70pt} \textbf{-l} \hspace{10pt} -ø}\\
      \gloss{AJ.preceding\_higher \textbf{+CA} +NOM}\\
      `the first' -- `el primer (puesto/lugar), el primero' -- `the first (position/place)'
\item \label{e5d} \mapuex{tripa \hspace{54pt} \textbf{-l} \hspace{10pt} -ø}\\
      \gloss{IV.go-out\_leave \textbf{+CA} +NOM}\\
      `end' -- `el [que ha] salido', `el de fuera' -- `the one [who has] gone out', `the one [from] outside'
\item \label{e5e} \mapuex{montu \hspace{54pt} \textbf{-l} \hspace{10pt} -ø}\\
      \gloss{IV.escape\_survive \textbf{+CA} +NOM}\\
      `protector' -- `el libertador', `el salvador' -- `the liberator', `the saviour'
\end{enumerate}

This analysis reflects a possible strategy available in \mapu{Mapudüngun} for creating nouns or adjectives (depending on the use given), from intransitive verbs or other verbalised categories. The causative is added to the verb stem, and the form is left uninflected and undeclined. For the purposes of the analysis, and due to the existence of nominalisers that function similarly but have a specific form, this mechanism is represented by the addition of a vacuous nominalizer, \mapu{-ø}. Finally, the nominalizer suffix \mapu{-fal} may not exist, but this is still an issue to be confirmed.

It should be noted, based on what has been discussed so far, that the restriction revealed by \citeauthor{smeets2008} (\citeyear{smeets2008}: 312)—that the suffix \mapu{-fal} only combines with transitive roots—can be reformulated. It can be proposed that transitive roots can be accompanied by the causative morpheme \mapu{-(ü)l} as long as they form a compound with the determiner \mapu{fa-} `this, here', and that this compound ultimately becomes an adjective or noun, as the verbal form from which it originates undergoes no derivation or inflection process.

Another possibility is that the \mapu{-fal} recognised as a nominalizer by \citeauthor{smeets2008} (\citeyear{smeets2008}: 312) and as an adjectiviser in \mapu{Düngupeyüm} may actually be the same suffix also recognised by her as \textsc{FORCE} (force majeure), with only slight variations in intensity and consequently in function, depending on its position. Also note that \citeauthor{smeets2008} (\citeyear{smeets2008}: 272) indicates that the suffix \mapu{-fal} ``...either (1) implies that there is a necessity or obligation for the subject to perform the action, or (2) that the subject orders someone else to perform the action''. And the causative, as \citeauthor{zuniga2006} (\citeyear{zuniga2006}: 390) indicates, is a ``...verbal voice that shows that the subject of the verb does not perform the action but instead commands or assigns others to do it''. As seen, the semantic effect is comparable, if not the same.

In the explanations for \hyperref[e18]{E.18} and \hyperref[e19]{E.19}, the suffix \textsc{FORCE}, which has the same form, \mapu{-fal}, will be related to the combination of the determiner \mapu{fa-} and the causative \mapu{-(ü)l}, just as in the previous paragraphs when explaining the alternative analysis of the adjectiviser (\textsc{ADJDO}).

In conclusion, it is proposed that the suffix \mapu{-fal} might suggest that the subject of the sentence is either required or obligated to carry out an action, or that the subject instructs someone else to perform the action when it appears in an inflected verbal form. However, this sense of obligation (or strong likelihood) transfers to the action itself when the suffix appears in a form that remains uninflected.

In both cases, there is the exception that the causative \mapu{-(ü)l} can occur with transitive verbal roots. Since everything related to the suffix \mapu{-fal} has not yet been confirmed, the system retains the rules that recognise it as both an adjectiviser and \textsc{FORCE}, as well as the rules that separate its components and analyse them as proposed.

Returning to the topic at hand, and as previously mentioned, verifying forms that end in \mapu{-fal} is a reliable method for determining that the verbal root with this suffix is transitive. Unfortunately, cases are very sporadic, making this task rather unproductive.

\subsection{\label{sec.03.3} Causative Suffix \mapu{-(ü)m}}

The causative \mapu{-(ü)m} triggers a valency-increasing operation \citep{kulikov2013} indicating that the subject does not perform the action but instructs others to do so (\citealp{zuniga2006}: 390). Therefore, in \mapu{Mapudüngun}, this suffix modifies the verbal voice to indicate that the subject requests, orders, or prompts someone else to perform the action expressed by the verb; as a result, intransitive verbs are transitivised. With regard to this causative morpheme, authors state the following:

\begin{quote} \label{qt03smeets}
The suffix \mapu{-(ü)m} is not productive and only combines with about 35 roots. the causative morpheme \mapu{-(ü)m} is found exclusively with intransitive verbs (\citealp{smeets2008}: 299).
\end{quote}

\begin{quote} \label{qt04golluscio}
The suffix \mapu{-(ü)m} is used exclusively with intransitive verbs... (\citealp{golluscio2007}: 210).
\end{quote}

\begin{quote} \label{qt05longkon}
The \mapu{Mapuche} language has two causative morphemes used to alter the transitivity of verbs. The morpheme \mapu{-(ü)l} (see \S\ref{sec.03.4}) is employed with agentive intransitive verbs, while the morpheme \mapu{-m} is used with unaccusative intransitive verbs\footnote{Inergative (agentive) and unaccusative verbs are two types of intransitive verbs; in the words of the author herself, ``inergative verbs denote activities or processes that depend on the will of an agent'' (\citealp{longkon2011a}: 85) and ``unaccusative verbs are those whose syntactic subjects are their notional objects. Their semantic function is to denote an affected or unaffected theme'' (\citealp{longkon2011a}: 86).\vspace{0.2cm}}... (\citealp{longkon2011a}: 110).
\end{quote}

It appears, according to the three citations presented above, that there is a strong basis for recognising intransitive verbal roots accompanied by the affix \mapu{-(ü)m}. All three authors agree that only intransitive roots can be combined with the causative morpheme \mapu{-(ü)m}, which also has the effect of making transitive a form derived from an intransitive root. To verify this assumption, examples from \cite{smeets2008} and some extracted from \mapu{Koña}'s autobiography \citep{mosbach1930} will be presented and analysed. Although the work of \cite{longkon2011a} has been consulted, material from this source is not included because the linguist does not delve deeply into this topic and provides very few examples. Examples from \cite{golluscio2007} are also not included because, after checking them out, of the 17 roots she presents as causativisable with \mapu{-(ü)m}, ten are intransitive and the remaining seven belong to other lexical categories, meaning they are not verbal roots according to the categorisation used in this work (see \citealp{chandia2025}); in summary, the author confirms her assertion with these examples, although only ten correspond to verbal roots.

With regard to the examples of \cite{smeets2008}, it has also been verified whether the causative suffix \mapu{-(ü)m} appears exclusively with intransitive verbal roots. Here, some of these examples will be presented, along with other instances where the author shows occurrences of a transitive root combined with the causative suffix \mapu{-(ü)m}. These cases will be explained, including some compounds with transitive roots that incorporate the suffix in question.

The suffix \mapu{-(ü)m} is the only one that causes a phonological change in the root that precedes it. This radical change is the rule and occurs as follows:

\paragraph{\example{6} \label{e6}}~
\vspace{-10pt}
\begin{enumerate}[label=\alph*.]
\item \label{e6a} \mapuex{la} `dead' + \mapuex{-üm} → \mapuex{la\textbf{ng}üm} `kill' (prothesis, sound insertion \mapuex{\textbf{ng}}\footnote{The digraph \mapu{\textbf{ng}} represents the /\textipa{\ng}/ sound in \mapu{Mapudüngun}.\vspace{0.2cm}})
\item \label{e6b} \mapuex{a\textbf{f}} `end' + \mapuex{-üm} → \mapuex{a\textbf{p}üm} `finish, complete' (alternation (`sandhi\footnote{Although the term `sandhi' is typically used for phonological changes occurring at word boundaries, it could also be applied here because the phonological changes occur at the morphemic boundaries of a polysynthetic language—one in which morphemes often correspond to entire words in other languages.\vspace{0.2cm}}') of \mapuex{\textbf{f}} for \mapuex{\textbf{p}})
\item \label{e6c} \mapuex{na\textbf{g}} `down' + \mapuex{-üm} → \mapuex{na\textbf{k}üm} `take down' (alternation (`sandhi') of \mapuex{\textbf{g}} for \mapuex{\textbf{k}})
\end{enumerate}

The following \hyperref[e7a]{E.7a} and \hyperref[e7c]{E.7c} illustrate two exceptions to the radical change just described, as shown by \cite{smeets2008}. However, these exceptions have not been found in the memoirs of \mapu{Paskwal Koña} \citep{mosbach1930}, although they do appear in de Augusta's dictionary \citep{chandia2014}, as shown in \hyperref[e7b]{E.7b} and \hyperref[e7d]{E.7d}.

\paragraph{\example{7} \label{e7}}~
\vspace{-10pt}
\begin{enumerate}[label=\alph*.]
\item \label{e7a} \mapuex{lle\textbf{g}üm-} `grow sth.' -- `cultivar algo' \hfill (\citealp{smeets2008}: 29)\\
\mapuex{lle\textbf{g}üm} `make come up (eg. plants)' \hfill (p. 53)\\
\mapuex{lle\textbf{g}-üm} `cause to come up (eg. plants)' \hfill (p. 299)\\
\mapuex{lle\textbf{g}-üm} `(Vt) make seeds come up (the sun)' \hfill (p. 528)
\item \label{e7b} \mapuex{lle\textbf{g}ümün}: (from \mapu{llegün}) tr. give birth, deliver (a baby) \hfill (de Augusta)
\item \label{e7c} \mapuex{na\textbf{g}üm} `cause to go down' \hfill (\citealp{smeets2008}: 29)\\
\mapuex{na\textbf{g}üm} `take down' \hfill (p. 53)
\item \label{e7d} \mapuex{na\textbf{g}ümün}: tr. Huapi\footnote{de Augusta refers to a topolect; \mapu{wapi} means `island'. Lake \mapu{Rangko}, located in the Los Ríos Region of Chile, has thirteen islands, the largest of which is known as ``Isla \mapu{Wapi}''; this is the area to which de Augusta refers. Meanwhile, in Argentine Patagonia, there is a lake called ``\mapu{Nawel Wapi}'' (Jaguar Island), with its largest island also named ``\mapu{Nawel Wapi}'', sometimes referred to simply as \mapu{Wapi}.\vspace{0.2cm}}. → \mapuex{na\textbf{k}ümün} \hfill (de Augusta)
\end{enumerate}

As for other examples from \cite{smeets2008}, \hyperref[e8a]{E.8a} provides the definition of \mapu{püna-}; it is the entry in her dictionary (p. 551) where this root is established as transitive. \hyperref[e8b]{E.8b} shows the subentry that includes the causative suffix (\textsc{CA}), which makes the verb transitive. If the root were indeed transitive, this would be an exception, or the verb might be misclassified in terms of valency. It has been confirmed that \mapu{püna-} is an intransitive verbal root, and therefore, it is possible to causativise this root by adding the morpheme \mapu{-(ü)m}. de Augusta lists \mapu{püna-} as intransitive, and \hyperref[e8c]{E.8c} to \hyperref[e8h]{E.8h} are all the verb forms generated with this verb found in the memoirs of \mapu{Paskwal Koña} \citep{mosbach1930}, all of which are used in an intransitive manner:

\paragraph{\example{8} \label{e8}}~
\vspace{-10pt}
\begin{enumerate}[label=\alph*.]
\item \label{e8a} \mapuex{püna-} (Vt) to stick, to glue -- `pegar, encolar'; \mapuex{pünaeyew ti chiklet} `she got stuck to her chewing gum' -- `ella se quedó pegada a su chicle'; \textit{literal:} `it stuck her gum'; \mapuex{pünafiñ} `I stuck to her, I hung around her' -- `me pegué a ella, me arrimé a ella'; \textit{literal:} `I stuck myself to him/her -- *me le pegué\footnote{\label{altisonante}Certainly, this is an ungrammatical (and jarring) expression in Spanish, but it best interprets the \mapu{Mapuche} verb in the example.\vspace{0.2cm}}'
\item \label{e8b} \mapuex{püna-m} (Vt) to glue something to -– `encolar/pegar algo a (algo más)'
\item \label{e8c} \mapuex{epu püna-le-ka-rke-fu-y} `two [of them] remained firmly stuck together'
\item \label{e8d} \mapuex{kom püna-wi-ng-ün} `all of them stuck together'
\item \label{e8e} \mapuex{kura mew püna-ntüku-le-y} `on the rocks, they are stuck'
\item \label{e8f} \mapuex{püna-kon-küle-y} `they get nailed'; \textit{literal:} `they are stuck entering'
\item \label{e8g} \mapuex{püna-n\footnote{\label{n.epent}The previous \hyperref[e8f]{E.8f} and the following \hyperref[e8h]{E.8h} suggest that this \mapu{-n} might be an epenthetic phonological element (a phonological adjustment) as in \hyperref[e8h]{E.8h} (\mapu{tüku- → ntüku-}) and not the Simple Verbal Noun morpheme as proposed by \cite{smeets2008} for some analyses of similar forms.\vspace{0.2cm}}-kon-küle-y} `they are found stuck'
\item \label{e8h} \mapuex{püna-ntüku-y trawa mew} `they stick strongly to the skin'
\end{enumerate}

The following examples illustrate some complexities of \mapu{Mapudüngun}. In \hyperref[e9a]{E.9a}, to the left of the label \textsc{CR.TV} (Transitive Compounded Stem) are the forms making up the verbal compound: two verbal roots (one intransitive, one transitive) and the causative \mapu{-(ü)m}. This example shows that the causative is added to the intransitive root when it appears first and the transitive root second, determining the compound's valency\footnote{The resulting verbal compound (from two or more verbal roots) derives its valency from the later component (\citealp{smeets2008}: 315).\vspace{0.2cm}}:

\paragraph{\example{9} \label{e9}}~
\vspace{-10pt}
\begin{enumerate}[label=\alph*.]
\item \label{e9a} \mapuex{püra \hspace{18pt} \textbf{-m} \hspace{16pt} -ye \hspace{50pt} -nie \hspace{8pt} -fi \hspace{8pt} -y \hspace{8pt} -ø \hspace{4pt} -iñ} \hfill (\citealp{smeets2008}: 474; 47)\\
      \gloss{IV.go-up \textbf{+CA} +TV.carry -CR.TV +PRPS +3P +IND +1 +PL}\\
      `we kept it up' -- `lo mantuvimos arriba/levantado/elevado'; \textit{literal:} `we made it go up by carrying it'
\item \label{e9b} \mapuex{tofkü \hspace{2pt} -ñ\footnote{The \mapu{-ñ-} between \mapu{tofkü-} and \mapu{püra-} is a phonological adjustment. In fact, \mapu{püra-}, which causes the appearance of this phonological prosthesis, should be represented as \mapu{ñpüra-}; see footnote \ref{n.epent}.\vspace{0.2cm}} \hspace{2pt} -püra \hspace{4pt} \textbf{-m}} \hfill (\citealp{smeets2008}: 558)\\
      \gloss{IV.spit +IV.go-up \textbf{+CA} -CR.IV}\\
      `(Vt) spit up [to]' -- `escupir hacia arriba [a]'; \textit{literal:} `spit going up [towards]'
\item \label{e9c} \mapuex{are \hspace{34pt} -ngül \hspace{12pt} \textbf{-üm} \hspace{20pt} -üñma \hspace{2pt} -e \hspace{10pt} -n \hspace{22pt} -ew \hspace{6pt} ñi \hspace{6pt} mansun} \hfill (\citealp{smeets2008}: 227; 10)\\
      \gloss{TV.lend +IV.gather \textbf{+CA} -CR.TV +IO +INV +IND1SG +3A SP.my NN.ox}\\
      `he lent my ox' -- `él prestó mi buey'; \textit{literal:} `he lent [it] making [it] join [with the others] my ox'
\item \label{e9d} \mapuex{reng \hspace{30pt} \textbf{-üm} \hspace{10pt} -nak \hspace{10pt} \textbf{-üm}} \hfill (\citealp{smeets2008}: 554)\\
      \gloss{IV.sediment \textbf{+CA} +AV.down \textbf{+CA} -CR.TV}\\
      `(Vt) cause something to settle and thicken' -- `causar que algo se asiente y espese';\\
      \textit{literal:} `make [it] sediment by lowering [it]'\footnote{Including these examples has revealed that, although combining the verbal roots would result in an intransitive verbal compound, the addition of the causative to the second root causes the compound to become transitive. This observation will be added as a rule to the \mapu{Düngupeyüm}.\vspace{0.2cm}}
\end{enumerate}

When the compound is formed by two intransitive roots, as in \hyperref[e9b]{E.9b}, the causative can be attached to either of the two roots (in this case, the second one) within the verbal compound. Conversely, when the first root is transitive and the second is intransitive, as in \hyperref[e9c]{E.9c} (reverse order compared to \hyperref[e9a]{E.9a}), the causative suffix \mapu{-(ü)m} is added to the second root, which remains within the verbal compound.

From this perspective, \hyperref[e9d]{E.9d} confirms the method for forming compounds as detailed above. Here, there is a double causativisation; reviewing the literal translation reveals a kind of pleonasm—two combined actions to achieve the goal indicated by the compound verbal form. Additionally, as observed in \hyperref[e9d]{E.9d} about verb + verb compounds, ``both the first and the second verbal roots in such a compound can take a suffix from slots 33 to 35\footnote{The suffixes from these slots are: \mapu{-ka} Factitive: S33, \mapu{-tu} Applicative: S33, \mapu{-(ü)l} Causative: S34, \mapu{-(ü)m} Causative: S34, \mapu{-ye} Oblique Object: S34, \mapu{-(ñ)ma} Experimentative: S35.\vspace{0.2cm}}'' (\citealp{smeets2008}: 315).

The following examples come from the autobiography of the \mapu{longko Paskwal Koña} \citep{mosbach1930}, and they confirm, as indicated by the three cited authors regarding this topic, that the causative morpheme \mapu{-(ü)m} only combines with intransitive verbal roots (\hyperref[e10a]{E.10a}). However, it also combines with nominal roots (\hyperref[e10b]{E.10b}), adjectival roots (\hyperref[e10c]{E.10c}), adverbial roots (\hyperref[e10d]{E.10d}), and a couple of demonstratives (\hyperref[e10e]{E.10e}):

\paragraph{\example{10} \label{e10}}~
\vspace{-10pt}
\begin{enumerate}[label=\alph*.]
\item \label{e10a} \mapuex{füy \hspace{70pt} \textbf{-üm} \hspace{4pt} -el}\\
      \gloss{IV.tighten\_hold-on \textbf{+CA} +OVN}\\
      `held, fastened'; \textit{literal:} `make it held/fastened'
\item \label{e10b} \mapuex{chaw \hspace{44pt} \textbf{-üm} \hspace{2pt} -ke \hspace{10pt} -i \hspace{10pt} -ng \hspace{2pt} -u}\\
      \gloss{NN.father\_man \textbf{+CA} +HAB +IND +3 +DL}\\
      `they brood'; \textit{literal:} `they make fathering\footnote{The \mapu{Mapuche} use the term `fathering' to refer to `incubating eggs'. The habituative suffix \mapu{-ke} can either indicate that it is a habitual action of birds or that the action extends over time as a custom. Finally, the dual number suffix \mapu{-u} might have been used mistakenly instead of \mapu{-ün}, which indicates an indeterminate plural, or it may refer to the two entities involved in the action, the bird and the egg.\vspace{0.2cm}}'
\item \label{e10c} \mapuex{firkü \hspace{46pt} \textbf{-m} \hspace{6pt} -nge \hspace{4pt} -ke \hspace{10pt} -y \hspace{8pt} -ø}\\
      \gloss{AJ.fresh\_cool \textbf{+CA} +PASS +HAB +IND +3}\\
      `they let [it] cool down'; \textit{literal:} `they make [it] be cool'
\item \label{e10d} \mapuex{üng \hspace{20pt} \textbf{-üm} \hspace{2pt} -uw \hspace{6pt} -pu \hspace{8pt} -y \hspace{10pt} -ø \hspace{2pt} -iñ}\\
      \gloss{AV.while \textbf{+CA} +REF +LOC +IND +1 +PL}\\
      `we waited there (the rest of us)'; \textit{literal:} `we made ourselves whiling there'
\item \label{e10e} \mapuex{fa \hspace{48pt} \textbf{-m} \hspace{6pt} -a \hspace{12pt} -el}\\
      \gloss{DP.this\_here \textbf{+CA} +FUT +OVN}\\
      `do this'
\end{enumerate}

The list of verbs identified by the authors as limited has expanded upon reviewing the text dictated by the \mapu{longko} to the Capuchin priest \citep{mosbach1930}. In Table 4 of \cite{chandia2025} are all the roots that have so far been confirmed to combine with the causative suffix \mapu{-(ü)m}; this list is not definitive as there are still uncertainties in several cases, and additionally, there are 545 more words yet to be verified.

To conclude the review of the causative suffix \mapu{-(ü)m}, it has been observed that it is a reliable mechanism for determining that the verb root it attaches to is intransitive. However, it is also necessary to first verify whether the root is indeed verbal or belongs to another category. Furthermore, as noted by the authors, not all intransitive roots accept causativisation with \mapu{-(ü)m}. In this regard, it would be important to assess the claims made by the three authors concerning the role of telicity or atelicity of verbs in the selection of the causative suffix.

\subsection{\label{sec.03.4} Causative Suffix \mapu{-(ü)l}}

According to \cite{smeets2008}, the allomorph of \mapu{-l} is \mapu{-ül}, whereas for \cite{golluscio2007} it is \mapu{-el}. \citeauthor{smeets2008} (\citeyear{smeets2008}: 300) adds that ``the suffix \mapu{-ül} has the optional allomorph \mapu{-el}'' for some verbs. The interchange between \mapu{ü} and \mapu{e} is common in \mapu{Mapudüngun}, often even within the same speaker. In the pair of articles being presented, we use \mapu{-ül} (hence we present the morpheme as \mapu{-(ü)l}), as it appears to be the preferred form in the \mapu{Ngülumapu} dialects, the \mapu{Mapuche} territory extending across present-day Chile.

\cite{longkon2011a} has already been cited regarding this suffix. According to this citation, she notes that the suffix \mapu{-(ü)l} is also causative and is added to intransitive verb roots, specifically those of the inergative type.

Regarding this morpheme, the other two authors state that:

\begin{quote} \label{qt06golluscio}
``\mapu{-(ü)l} may derive from the verb \mapu{el-}, which means `place [in the world], create, form'\footnote{It also means `leave', as in `leave something placed', `leave something forgotten', `leave something painted', etc. (see footnote \ref{kununote}).\vspace{0.2cm}}, and \mapu{-(ü)l} is used with intransitive and labile verbs'' (see \S\ref{sec.03.4.1}). Note\footnote{This text is included as a footnote in the cited text; it has been deemed appropriate to reproduce it here as well.\vspace{0.2cm}}: Both Smeets (1989) and Aranovich (2003) provide examples of causativisation of transitive verbs with \mapu{-(ü)l}. However, my consultant Fresia \mapu{Melliko} mentioned that the basic function of \mapu{-(ü)l} in these verbal constructions is applicative; the causative interpretation is possible, but depends on the discourse context'' (\citealp{golluscio2007}: 209).
\end{quote}

\begin{quote} \label{qt07smeets}
``The suffix \mapu{-(ü)l} is productive. It combines with both transitive and intransitive verbs, as well as with loan verbs (unlike \mapu{-(ü)m})'' (\citealp{smeets2008}: 299).
\end{quote}

The three authors agree that the causative suffix \mapu{-(ü)l} combines with intransitive roots, whether they are of \mapu{Mapuche} origin or Spanish loanwords, which is not the case for the causative \mapu{-(ü)m}, which only combines with \mapu{Mapudüngun} roots. However, \cite{longkon2011a} argues that the \textsc{CA} suffix only combines with intransitive roots, while \cite{golluscio2007} slightly expands the suffix's applicability to include labile verbal roots (see \S\ref{sec.03.4.1}). On the other hand, \cite{smeets2008} directly includes transitive roots among the forms that can present the morpheme \mapu{-(ü)l}\footnote{\label{ca-mio-ben}\cite{smeets2008} reports three very similar suffixes in form and allomorphy, which also occur in adjacent slots: the `Causative' \mapu{-l} with allomorphs \mapu{-ül}, \mapu{-el} and \mapu{-lel}; slot 34. The `More Implicated Object' \mapu{-l} with allomorphs \mapu{-ül} and \mapu{-el}; slot 29. And the `Benefactive' \mapu{-el} with allomorphs \mapu{-l} and \mapu{-lel}; slot 27. \citeauthor{zuniga2009b} (\citeyear{zuniga2009b}: 18) comments that he has not found sufficient evidence to support the view that the benefactive is a distinct suffix separate from the causative. In support of Harmelink's conclusion, reformulated by \cite{zuniga2009b}---``there is only one suffix that not only causativises intransitives but also applicativises transitives''---the following observation is made: in \cite{smeets2008}, in all examples containing both suffixes, one is always followed by the other (\textsc{CA+BEN}, \mapu{-l-el}), meaning that no intermediate suffixes occur, even though six slots have been established between \textsc{CA} and \textsc{BEN} where other suffixes could appear. Slot 33: \mapu{-tu} `Applicative', \mapu{-ka} `Factitive'. Slot 32: \mapu{-nie} `Progressive Persistent', \mapu{-künu} `Perfect Persistent'. Slot 31: \mapu{-(u)w} `Reflexive/Reciprocal'. Slot 30: \mapu{-(k)iaw} `Circular Movement', \mapu{-(kü)tie} `Intensifier'. Slot 29: \mapu{-(ü)l} `More Implicated Object'. Slot 28: \mapu{-(kü)le} `Stative', \mapu{-meke} `Progressive'. It would be worth checking other texts to see if other morphemes occur between \mapu{-l} and \mapu{-(l)el}.\vspace{0.2cm}}.

\subsubsection{\label{sec.03.4.1} Labile Roots and the Causative \mapu{-(ü)l}}

Labile verbs in \mapu{Mapudüngun} are roots that can be used transitively or intransitively without needing specific suffixes to act transitively. \citeauthor{golluscio2007} (\citeyear{golluscio2007}: 214) provides two examples of labile roots combining with the causative suffix \mapu{-(ü)l}: \mapu{kim-} and \mapu{watro-}. However, in this study, the root \mapu{kim-} is excluded because, according to the categorical classification in \cite{chandia2025} (see \S\ref{sec.02.1}), it belongs to the adjective class (\textsc{AJ.\mapu{kim}\_wise\_knower}) and is thus not a true verbal root but a verbalised one, making it prone to either valency.

In the case of \mapu{watro-} `split', according to de Augusta's definition \citep{chandia2014} and the examples provided by \citeauthor{golluscio2007} (\citeyear{golluscio2007}: 214; 10a, 10b), it appears to be an ambitransitive\footnote{As \cite{golluscio2007} states, labile verbs are a subtype of ambitransitive verbs in which the same verb form is used for both transitive and intransitive meanings, but without any morphological change.\vspace{0.2cm}} verb. \mapu{Paskwal Koña} \citep{mosbach1930} uses this verb only four times (\hyperref[e11a]{E.11a} to \hyperref[e11d]{E.11d}), but none are in the transitive form:

\paragraph{\example{11} \label{e11}}~
\vspace{-10pt}
\begin{enumerate}[label=\alph*.]
\item \label{e11a} \mapuex{\textbf{watro} \hspace{12pt} -nge \hspace{6pt} -y \hspace{8pt} -ø \hspace{12pt} ñi \hspace{22pt} i \hspace{16pt} -nge \hspace{6pt} -a \hspace{12pt} -l}\\
      \gloss{IV.split +PASS +IND +3 SP.their TV.eat +PASS +FUT +OVN}\\
      `they are harvested for consumption'; \textit{literal:} `they are split for eating [later]'
\item \label{e11b} \mapuex{\textbf{watro} \hspace{12pt} -n \hspace{18pt} uwa \hspace{16pt} fün \hspace{10pt} -küle \hspace{2pt} -lu}\\
      \gloss{IV.split +PVN NN.corn NN.fruit +ST +SVN}\\
      `the spadix separated from the plant'; \textit{literal:} `[the] split [of the] maize [is what] is spiked'
\item \label{e11c} \mapuex{\textbf{watro} \hspace{18pt} -ntu \hspace{40pt} -ye \hspace{6pt} -a \hspace{12pt} -fi \hspace{8pt} -ñ \hspace{30pt} wün \hspace{16pt} mew}\\
      \gloss{IV.split -TV.take -CR.TV +OO +FUT +3P +IND1SG NN.mouth INST}\\
      `I will crush it between my teeth'; \textit{literal:} `I will take it to split in [my] mouth'
\item \label{e11d} \mapuex{fey \hspace{24pt} \textbf{watro} \hspace{8pt} -y \hspace{8pt} -ø \hspace{6pt} kuykuy}\\
      \gloss{AV.then IV.split +IND +3 NN.bridge}\\
      `that [one] broke'; \textit{literal:} `then it split [the] bridge\footnote{This phrase comes from an \mapu{epew} (a type of fable) told by \mapu{Paskwal Koña}, where the `bridge' refers to a stick used as a makeshift crossing.\vspace{0.2cm}}'
\end{enumerate}

According to \citeauthor{smeets2008} (\citeyear{smeets2008}: 570), \mapu{watro-} is considered labile, meaning `(Vi, Vt) break'. However, only one of her analysed examples is intransitive (\hyperref[e12a]{E.12a}), which \cite{smeets2008} herself identifies as such. Another example (\hyperref[e12b]{E.12b}) shows \mapu{watro-} used transitively but no analysis is presented. In this case segmentation and analysis are done using \mapu{Düngupeyüm}. Without additional context, it's unclear if the translation is accurate. A translation suggesting an intransitive use could be `his cup broke completely'.

\paragraph{\example{12} \label{e12}}~
\vspace{-10pt}
\begin{enumerate}[label=\alph*.]
\item \label{e12a} \mapuex{watro \hspace{12pt} -ka \hspace{8pt} -w \hspace{8pt} -üy \hspace{8pt} -ø} \hfill (\citealp{smeets2008}: 388; 33)\\
      \gloss{IV.split +FAC +REF +IND +3}\\
      `it broke into (various) pieces; it was destroyed' -- `se quebró en (varios) pedazos; fue destruido'; \textit{literal:} `it broke completely'
\item \label{e12b} \mapuex{watro \hspace{44pt} -ka \hspace{8pt} -y \hspace{10pt} -ø \hspace{6pt} ñi \hspace{10pt} tasa} \hfill (\citealp{smeets2008}: 570)\\
      \gloss{IV.break\_split +FAC +IND +3 SP.his NN.cup}\\
      `he broke his cup in various pieces' -- `quebró su taza en varios pedazos'
\end{enumerate}

In \cite{smeets2008}, only two roots have been identified that, according to the categorisation in this work (see \S\ref{sec.02.1}), are verbal, combine with the causative suffix \mapu{-(ü)l}\footnote{According to \citeauthor{smeets2008} (\citeyear{smeets2008}: 300), \mapu{-el} is a less common allomorph that appears only with certain verbs.\vspace{0.2cm}}, are ambitransitive according to her notes, and provide analysed examples. Sometimes the author notes that certain roots combine with specific suffixes but, unfortunately, does not always provide examples. These roots are \mapu{llüka-} meaning `be frightened, fear,' and \mapu{üta-} meaning `graze.' For \mapu{üta-}, there is only one example (\hyperref[e13b]{E.13b}) that includes the causative suffix \mapu{-(ü)l}, and there isn't enough material in this work to determine the valency behaviour of the root. However, consulting CORLEXIM \citep{chandia2014}, both de Valdivia and Febrés indicate that the root corresponding to `graze' includes the causative, \mapu{üta-l-ün}, meaning `make graze'. de Augusta, on the other hand, does document the ditransitive use and provides an example of a transitive use (\hyperref[e13a]{E.13a}), which again could be translated intransitively as ``the horse grazes [on grass]'':

\paragraph{\example{13} \label{e13}}~
\vspace{-10pt}
\begin{enumerate}[label=\alph*.]
\item \label{e13a} \mapuex{üta-ke-y kachu kawellu} \hfill (de Augusta)\\
      `the horse eats [grazes] grass'
\item \label{e13b} \mapuex{üta \hspace{24pt} \textbf{-l} \hspace{8pt} -uw \hspace{4pt} -küle \hspace{2pt} -a \hspace{8pt} -fu \hspace{6pt} -y \hspace{8pt} -ø} \hfill (\citealp{smeets2008}: 453; 22)\\
      \gloss{IV.graze \textbf{+CA} +REF +ST +FUT +RI +IND +3}\\
      `they graze [their cattle]' -- `ellos apacentan [su ganado]'; \textit{literal:} `they could make graze [their livestock]'
\end{enumerate}

The \mapu{longko Koña} \citep{mosbach1930} forms only one\footnote{There is another verb with this root in \mapu{Koña} \citep{mosbach1930}: \mapu{\textbf{ütal}-tüku-lel-nge-ke-y} \textsc{TV.ütaf\_press\_fit-in -TV.tuku\_put\_insert -CR.TV +BEN +PASS +HAB +IND +3}. However, this corresponds to the meaning recorded by de Valdivia for a form differing by one phoneme, \mapu{ütaf-}: `press, fit in'. The phrase from the text is \mapu{Ka kiñe fücha kuchillo \textbf{ütal}tükulelngekey...} translated as `they fit a large knife', which suggests there might have been an error in transcription or a slip of the tongue by the narrator.\vspace{0.2cm}} verb with the root \mapu{üta-} and uses it intransitively:

\paragraph{\example{14} \label{e14}}~
\vspace{-10pt}
\begin{enumerate}[label=\alph*.]
\item[] \mapuex{kachu \hspace{20pt} ñi \hspace{16pt} üta \hspace{16pt} -ya \hspace{8pt} -m}\\
\gloss{NN.grass SP.its IV.graze +FUT +IVN}\\
`its grazing'; \textit{literal:} `grass for [its] grazing'
\end{enumerate}

The root \mapu{llüka-} `become afraid, fear' is not definitively shown to be labile based on the examples in \cite{smeets2008}. The root is described as `(Vi, Vt) become afraid, fear'. This definition does not inherently suggest transitivity in English or Spanish (`tener miedo, temer'). The book includes an example (\hyperref[e15a]{E.15a}) that might imply a transitive use of the verb with \mapu{llüka-}. However, this transitivity depends on the translation, and while `fear' or `temer' can be both transitive and intransitive, this does not guarantee the same usage in other languages, as seen in the two possible literal interpretations of \hyperref[e15a]{E.15a}.

Nonetheless, during the course of this study, the root \mapu{llüka-} reappeared while examining the lability of certain verbs that manifest their transitive sense. In these cases, it is possible to use person agreement suffixes, which are exclusively used with transitive forms (see \S\ref{sec.03.5}). In that section, the lability of this verbal root was indeed confirmed.

In the next examples, \hyperref[e15b]{E.15b} and \hyperref[e15c]{E.15c} illustrate how the causative suffix transitivises the verb final form. Finally, the demonstration is provided by \hyperref[e15d]{E.15d} and \hyperref[e15e]{E.15e} that the same verb roots maintain their intransitivity in complete verbal forms in the absence of valency-altering morphemes.

\paragraph{\example{15} \label{e15}}~
\vspace{-10pt}
\begin{enumerate}[label=\alph*.]
\item \label{e15a} \mapuex{llüka \hspace{82pt} -ya \hspace{10pt} -e \hspace{10pt} -y \hspace{8pt} -ø \hspace{4pt} -u \hspace{8pt} -ø} \hfill (\citealp{smeets2008}: 193; 57)\\
      \gloss{IV.become-afraid\_fear +FUT +INV +IND +1 +DL +1t2A}\\
      `you expect me to be afraid of you' -- `tu esperas que te tenga miedo';\\
      \textit{literal:} `I will fear you / I will be frightened with you'
\item \label{e15b} \mapuex{llüka \hspace{82pt} \textbf{-l} \hspace{8pt} -ka \hspace{8pt} -ke \hspace{8pt} -fu \hspace{4pt} -y \hspace{6pt} -ng \hspace{2pt} -ün} \hfill (\citealp{smeets2008}: 228; 6)\\
      \gloss{IV.become-afraid\_fear \textbf{+CA} +FAC +HAB +RI +IND +3 +PL}\\
      `they used to intimidate' -- `ellos solían intimidar'; \textit{literal:} `they used to cause a lot of fear'
\item \label{e15c} \mapuex{llüka \hspace{82pt} \textbf{-l} \hspace{8pt} -ka \hspace{6pt} -che \hspace{28pt} -ke \hspace{10pt} -y \hspace{8pt} -ø} \hfill (\citealp{smeets2008}: 375; 25)\\
      \gloss{IV.become-afraid\_fear \textbf{+CA} +FAC +NN -CR.IV +HAB +IND +3}\\
      `he frightens people' -- `él asusta a las personas'; \textit{literal:} `he causes fear in people'
\item \label{e15d} \mapuex{llüka \hspace{84pt} -la \hspace{10pt} -y \hspace{8pt} -ø} \hfill (\citealp{smeets2008}: 420; 56)\\
      \gloss{IV.become-afraid\_fear +NEG +IND +3}\\
      `he was not afraid' -- `él no se asustaba; él no estaba atemorizado'; \textit{literal:} `he did not fear'
\item \label{e15e} \mapuex{llüka \hspace{82pt} -ke \hspace{8pt} -fu \hspace{6pt} -n} \hfill (\citealp{smeets2008}: 417; 28)\\
      \gloss{IV.become-afraid\_fear +HAB +RI +IND1SG}\\
      `I became afraid' -- `tuve miedo; me volví temeroso'; \textit{literal:} `I would always fear'
\end{enumerate}

In the autobiography of the \mapu{longko} \citep{mosbach1930}, twenty-one verbal roots\footnote{About one hundred and ten verbal roots that were not forming compounds were checked. Since this task took too much time, we decided to postpone the verification of these roots when they participate in verbal compounds and exhibit a suffix in one of the forms that the causative can take: \mapu{-l, -ül, -el}.\vspace{0.2cm}} that combine with the causative suffix \mapu{-(ü)l} were found. Almost all the roots are intransitive, except for two that form both transitive and intransitive verbs, meaning they are labile or bivalent roots, as shown in \hyperref[e16]{E.16} below.

\paragraph{\example{16} \label{e16}}~
\vspace{-10pt}
\begin{enumerate}[label=\alph*.]
\item \label{e16a} \mapuex{fey \hspace{18pt} mew \hspace{6pt} wüla \hspace{12pt} küme \hspace{6pt} monge \hspace{4pt} -ke \hspace{10pt} -ø \hspace{8pt} -ø \hspace{4pt} -iñ}\\
      \gloss{DP.that INST AV.until AJ.good IV.live +HAB +IND +1 +PL}\\
      `then we really came back to life'; \textit{literal:} `until that, we [didn't] live well'
\item \label{e16b} \mapuex{monge \hspace{6pt} -ke \hspace{8pt} \textbf{-fi} \hspace{8pt} -ñ \hspace{36pt} fill \hspace{34pt} lawen \hspace{12pt} mew}\\
      \gloss{IV.live +HAB \textbf{+3P} +IND1SG AJ.complete NN.medicine INST}\\
      `I administer various medications to them'; \textit{literal:} `I revive them with all kinds of medicine'
\item \label{e16c} \mapuex{monge \hspace{8pt} \textbf{-l} \hspace{10pt} -ke \hspace{8pt} -fi \hspace{8pt} -ø \hspace{8pt} -ø \hspace{2pt} -iñ \hspace{10pt} kutran \hspace{10pt} -chi \hspace{10pt} che}\\
      \gloss{IV.heal \textbf{+CA} +HAB +3P +IND +1 +PL NN.illness +ADJ NN.person}\\
      `we heal the sick'; \textit{literal:} `we make heal sick people'
\item \label{e16d} \mapuex{la \hspace{28pt} -ya \hspace{8pt} -lu \hspace{24pt} ka, \hspace{28pt} monge \hspace{12pt} -ke \hspace{8pt} -fi \hspace{8pt} -ø \hspace{8pt} -ø \hspace{2pt} -iñ}\\
      \gloss{AJ.dead +FUT +SVN CJ.AND (also) TV.revive +HAB +3P +IND +1 +PL}\\
      `even if they are already dying, we bring them back to life'; \textit{literal:} `[those who are] dying, we also bring back to life'
\item \label{e16e} \mapuex{yewe \hspace{42pt} -y \hspace{10pt} -ø \hspace{6pt} küpa \hspace{10pt} -ya \hspace{10pt} -el}\\
      \gloss{IV.be-ashamed +IND +3 IV.come +FUT +OVN}\\
      `he is embarrassed to come in'; \textit{literal:} `he feels ashamed of coming in'
\item \label{e16f} \mapuex{küme \hspace{14pt} may \hspace{26pt} poye \hspace{22pt} -w \hspace{10pt} -a \hspace{14pt} -i \hspace{10pt} -y \hspace{4pt} -u \hspace{8pt} ka \hspace{22pt} yewe \hspace{16pt} -w \hspace{12pt} -a \hspace{12pt} -i \hspace{12pt} -y \hspace{4pt} -u}\\
      \gloss{AJ.good CJ.yes TV.appreciate +REF +FUT +IND +1 +DL CJ.AND TV.respect +REF +FUT +IND +1 +DL}\\
      `we will esteem and respect each other'; \textit{literal:} `yes indeed, we will esteem and respect each other'
\item \label{e16g} \mapuex{yewe \hspace{42pt} \textbf{-l} \hspace{8pt} -ka \hspace{10pt} -ntüku \hspace{8pt} -ke \hspace{8pt} -e \hspace{10pt} -l \hspace{10pt} -i \hspace{4pt} -ø \hspace{10pt} -ø}\\
      \gloss{IV.be-ashamed \textbf{+CA} +FAC +TV.put +NEG +INV +SJI +1 +SG +1t2A}\\
      `don't make me feel ashamed'; \textit{literal:} `If (truly) you did not put me in a situation that causes me embarrassment'
\end{enumerate}

In \hyperref[e16b]{E.16b}, where the verb conveys a transitive meaning, the morpheme \mapu{-fi} is used. This suffix marks a third-person patient (Differential Object Marker, \textsc{DOM}), and may appear whether or not the patient is explicitly mentioned in the phrase. \mapu{Mapudüngun} also has an inverse form\footnote{\label{dir-inv} Most linguistic researchers of \mapu{Mapudüngun} define the verbal system in this way (direct, inverse, and passive). This system is explored in detail by \cite{golluscio2010}, \cite{salas1992a} and \cite{zuniga2006}.\vspace{0.2cm}} which uses the morpheme \mapu{-e} instead (see the discussion of person agreement suffixes in \S\ref{sec.03.5}). The suffix \mapu{-fi} does not indicate number or gender; these aspects can sometimes be inferred from the sentence context.

\cite{smeets2008} referred to \mapu{-fi} as an `External Direct Object' (\textsc{EDO}), but subsequent research has replaced this term with \textsc{DOM} (Differential Object Marker) for third-person patients, and the marker previously called `Internal Direct Object' is now understood as an inverse marker (\textsc{INV}). These suffixes are best understood as person agreement markers that help track the participant roles in transitive events (see \S\ref{sec.03.5}). 

Conversely, in \hyperref[e16a]{E.16a}, neither \mapu{-fi} nor \mapu{-e} can appear, as the verb form is intransitive. This contrast illustrates how person agreement suffixes are restricted to transitive constructions\footnote{In an initial misinterpretation, it was thought that these morphemes also had the property of transitivising intransitive verbal roots. Thanks are due to Fernando Zúñiga, who pointed out this error and highlighted the labile nature of these verbal roots.\vspace{0.2cm}}.

In \hyperref[e16c]{E.16c} and \hyperref[e16d]{E.16d}, note that the only difference between the verbs formed from the root \mapu{monge-} `live, revive, heal' is the presence of the causative suffix \mapu{-(ü)l} in \hyperref[e16c]{E.16c}. These two verb forms are found in the same paragraph of the \mapu{longko}'s memoirs \citep{mosbach1930}, in fact, in two consecutive sentences. In this work, the verb corresponding to `heal' always appears with the causative, meaning that the root, in its intransitive use `heal', is transitivised by the causative suffix; `make heal' would be a more accurate translation. On the other hand, when the translation leans more towards the sense of `revive (someone)', the causative suffix does not appear in the verb form, indicating that the verb is being used transitively:

\begin{quote} \label{qt08kona}
\mapu{Fentenchi pu konamachi küdawtupelu, fey mew \textbf{monge-l-ke-fi-iñ} kutranchi che. Layalu ka, \textbf{monge-ke-fi-iñ}; taiñ adkünoetew taiñ chaw, fey mew femmekeiñ}.\\

`Since many \mapu{machi} assistants work alongside us, we heal the sick; even those who are already near death, we bring them back to life, with the empowerment given to us by our father god'.\\

\textit{literal:} `Many are the \mapu{machi}-servants who work close [to us], so we make them heal sick people. [Those who are] dying we also revive; [This is] our custom given [by] our father, so this is what we are doing'.\\
\end{quote}

In \hyperref[e16e]{E.16e} and \hyperref[e16f]{E.16f}, the verb \mapu{yewe-} is translated as `be embarrassed' when used in intransitive forms, and as `respect' when used transitively. However, in \mapu{Mapudüngun}, it is a single verb. In this and many other cases, we think that the semantics of this language should be studied from the \mapu{Mapuche} world-view to understand, if not fully, at least more accurately what a \mapu{Mapuche} means in their own language. It seems there may be a connection between the verb \mapu{ye-} `carry', the morpheme indicating interpersonal relation \mapu{-we-n}, and the verb \mapu{yewe-} `respect'. Note that to express a relationship with someone in \mapu{Mapudüngun}, a compound is formed with the verb \mapu{ye-}. For example, to say `he is my grandfather', the expression \mapu{laku-\textbf{ye-}fi-ñ} is used, which literally means `I carry him as grandfather' (\citealp{becerra2011}: 115). Also note that in the ``Chapter XI, Domestic Life'' of the \mapu{longko}'s memoirs \citep{mosbach1930}, when discussing family relationships, the verb \mapu{yewe-} `respect' and the relational suffix \mapu{-wen} are frequently used. Among other forms, you find \mapu{yewe-w-ün} `they respect each other', \mapu{llalla-we-n} `mother-in-law and son-in-law', \mapu{ewküll-we-n} `co-fathers-in-law', and \mapu{yewe-n wentru} `respectful men'. For instance, co-fathers-in-law refer to each other as \mapu{ewküll-we-n} and/or \mapu{yewe-n wentru}.

Regarding the combinability of the verbal root \mapu{yewe-} with the causative \mapu{-(ü)l}, only the form found in \hyperref[e16g]{E.16g} was discovered in \mapu{Koña}'s memoirs \citep{mosbach1930}. This example also shows that the causative combines with intransitive roots or with the intransitive sense of labile roots. However, since this is only one example, it cannot be conclusively stated that this rule is confirmed.

\subsubsection{\label{sec.03.4.2} Considerations on Certain Verbal Roots}

This section examines three cases in which roots or suffixes give rise to certain observations concerning the element emphasised in the respective explanations. All three cases present the causative morpheme \mapu{-(ü)l} or one of its various forms, and their intricacies require a thorough examination to elucidate their implications for the verb form, the meaning, and ultimately, the valency value.

\paragraph*{\label{sec.03.4.2.1}3.4.2.1\quad\mapu{troki-} `separate, set apart'} 

Among the verbal roots in \mapu{Paskwal Koña}'s records \citep{mosbach1930} that combine with an allomorph of the \mapu{-l} morpheme is the transitive root \mapu{troki-}. The term is defined in a variety of sources as signifying `consider, think that, believe that' (\citealp{smeets2008}: 564); `indicate' (de Augusta and Febrés in \citealp{chandia2014}); and `division, portion, part' (de Augusta in \citealp{chandia2014}). While this isn't the place for a detailed explanation of the proposed definition, examples and definitions suggest that \mapu{troki-} can mean `setting something aside for consideration'. The verbal form prompting the examination of \mapu{troki-} is shown in the following example:

\paragraph{\example{17} \label{e17}}~
\vspace{-10pt}
\begin{enumerate}[label=\alph*.]
\item[] \mapuex{ñuke \hspace{42pt} -ye \hspace{20pt} -l \hspace{16pt} troki \hspace{10pt} \textbf{-l} \hspace{12pt} -ke \hspace{10pt} -e \hspace{10pt} -y \hspace{10pt} -ø \hspace{4pt} -ew \hspace{10pt} ñi \hspace{16pt} pu \hspace{8pt} yall}\\
\gloss{NN.mother +TV.carry +OVN TV.split \textbf{+BEN} +HAB +INV +IND +3 +3A SP.their COLL NN.son}\\
`receives treatment from them as a stepmother';\\
\textit{literal:} `she's treated as a mother; they consider [like] that, their children (father's)'
\end{enumerate}

The morphological analysis system yielded the result shown in \hyperref[e17]{E.17}, where it makes considerable sense for the morpheme \mapu{-l} to correspond to a benefactive, as indicated by the literal translation of the example (see also footnote \ref{ca-mio-ben}). No conclusion is presented here regarding the morpheme and the semantic-grammatical value it contributes to the verbal form, but these details are noted and may be relevant for any research requiring such data.

\paragraph*{\label{sec.03.4.2.2}3.4.2.2\quad\mapu{pi-} `say'}

Moving on to another verbal root, according to some analyses with \mapu{Düngupeyüm} \citep{chandia2021}, the verb \mapu{pi-}, when combined with the determiner \mapu{fa-}, could potentially combine with the causative \mapu{-(ü)l}. For instance, the verb \mapu{pifaleymün} could be analysed in three possible ways:

\paragraph{\example{18} \label{e18}}~
\vspace{-10pt}
\begin{enumerate}[label=\alph*.]
\item \label{e18a} \mapuex{pi \hspace{22pt} -fa \hspace{52pt} -le \hspace{6pt} -y \hspace{8pt} -m \hspace{2pt} -ün}\\
      \gloss{TV.say +DP.this -CR.TV +ST +IND +2 +PL}
\item \label{e18b} \mapuex{pi \hspace{22pt} -fal \hspace{18pt} -e \hspace{10pt} -y \hspace{8pt} -m \hspace{2pt} -ün \hspace{4pt} -ø}\\
      \gloss{TV.say +FORCE +INV +IND +2 +PL +1t2A}
\item \label{e18c} \mapuex{pi \hspace{22pt} -fa \hspace{52pt} -l \hspace{12pt} -e \hspace{8pt} -y \hspace{10pt} -m \hspace{2pt} -ün \hspace{4pt} -ø}\\
      \gloss{TV.say +DP.this -CR.TV +CA +INV +IND +2 +PL +1t2A}
\end{enumerate}

In the context where the phrase \mapu{pifaleymün Elewterio} is translated as `Eleuterio sends me to tell you' \citep{mosbach1930}, the analysis including the stative suffix (\textsc{ST}) (\hyperref[e18a]{E.18a}) is dismissed because it loses the sense of `send to'; with this morpheme (\textsc{ST}), the translation would correspond to `he is telling you' or `he tells you'. According to \citeauthor{smeets2008} (\citeyear{smeets2008}: 169), the stative function indicates a state that may or may not imply agentivity of the subject, providing examples like \mapu{amu-le-n} `I am going' -- `voy/estoy yendo' and \mapu{ellka-le-y} `it is hidden' -- `está escondido/oculto'. Additionally, with some verbs, \mapu{-(kü)le} can denote an ongoing event or a resultant state, as in \mapu{la-le-y} (\textsc{die+ST+IND+3}), `he is dead' -- `está muerto', or `he is dying' -- `está muriendo', or `he has died' -- `ha muerto / murió'.

In the subsequent analyses, the sense of `sending to' is present. \hyperref[e18b]{E.18b} shows \mapu{pi-} and \mapu{fal-} \textsc{FORCE} combined; and \hyperref[e18c]{E.18c}, where \mapu{pi-} would co-occur with the demonstrative \mapu{fa-} followed by the causative \mapu{-(ü)l} (see \S\ref{sec.03.2}).

\citeauthor{smeets2008} (\citeyear{smeets2008}: 272) notes that the suffix \mapu{fal-} can indicate either that the subject of the sentence is required or obligated to perform an action, or that the subject is commanding someone else to perform the action. Regarding \mapu{fal-}, denoted as \textsc{DP.this\_here+CA}, this bimorphemic\footnote{Following \cite{zuniga2009b}, the term `bimorpheme' is used here to refer to a morpheme formed by the combination of two morphemes.\vspace{0.2cm}} construction can be interpreted as `cause this', in reference to the verb. See the following examples from \cite{smeets2008}; the analyses of \hyperref[e19a]{E.19a} and \hyperref[e19c]{E.19c} are by the author, while those of \hyperref[e19b]{E.19b} and \hyperref[e19d]{E.19d} are the alternative proposed analyses:

\paragraph{\example{19} \label{e19}}~
\vspace{-10pt}
\begin{enumerate}[label=\alph*.]
\item \label{e19a} \mapuex{ellka \hspace{16pt} -künu \hspace{4pt} -lel \hspace{8pt} \textbf{-fal} \hspace{8pt} -ye \hspace{8pt} -nge \hspace{6pt} -y \hspace{10pt} -ø} \hfill (\citealp{smeets2008}: 273; 3)\\
      \gloss{TV.hide +PFPS +BEN \textbf{+FORCE} +PLR +PASS +IND +3}\\
      `various [things] have to remain hidden from him' -- `varias cosas deben permanecer ocultas a él';\\
      \textit{literal:} `they must be kept hidden from him'
\item \label{e19b} \mapuex{ellka \hspace{16pt} -künu \hspace{4pt} -lel \hspace{14pt} \textbf{-fa \hspace{10pt} -l} \hspace{10pt} -ye \hspace{8pt} -nge \hspace{6pt} -y \hspace{8pt} -ø}\\
      \gloss{TV.hide +PFPS +BEN \textbf{+DP.this +CA} +PLR +PASS +IND +3}\\
      `they must be left hidden from him'
\item \label{e19c} \mapuex{lang\footnote{In \cite{smeets2008}'s analysis, the form \mapu{la-} corresponds to the verb `die', meaning that depending on its context, she recognises it as a verb, noun, or adjective. In \cite{chandia2025}, it has been determined that its basic form is adjectival. It could be argued that when \mapu{la-} is followed by \mapu{ng}, it is undeniably a verb, but it should be noted that \mapu{ng} is merely a phonological adjustment.\vspace{0.2cm}} \hspace{8pt} \textbf{-üm \hspace{4pt} -fal} \hspace{6pt} -ma \hspace{2pt} -e \hspace{10pt} -n \hspace{20pt} -ew} \hfill (\citealp{smeets2008}: 273; 6)\\
      \gloss{AJ.dead \textbf{+CA +FORCE} +IO +INV +IND1SG +3A}\\
      `he had it killed', `he has to kill it' -- `él lo mató', `él debe matarlo'; \textit{literal:} `he must kill it for me'
\item \label{e19d} \mapuex{lang \hspace{14pt} \textbf{-üm \hspace{8pt} -fa \hspace{12pt} -l} \hspace{8pt} -ma \hspace{2pt} -e \hspace{10pt} -n \hspace{20pt} -ew}\\
      \gloss{AJ.dead \textbf{+CA +DP.this +CA} +IO +INV +IND1SG +3A}\\
      `he has to kill it for me'
\end{enumerate}

In \hyperref[e19d]{E.19d}, by maintaining the analysis as \mapu{fa-l}, that is, the bimorphemic \textsc{DP+CA}, there is a double causativisation in the form, which is not unusual in \mapu{Mapudüngun}. The second causativisation acts upon a verb that has been transitivised by the first causative. This leads to two possible interpretations: firstly, the causative \mapu{-(ü)l} can indeed combine with transitivised (or transitively derived) roots, as reported by \cite{smeets2008} (see \hyperref[qt07smeets]{quote}, p. \pageref{qt07smeets}). Secondly, the causative \mapu{-(ü)l} appears because the main verbal root in the form is intransitive\footnote{As indicated by the analysis in \hyperref[e19d]{E.19d}, in reality, \mapu{la-} is an adjectival root that, when verbalised, becomes the intransitive verb, `die'.\vspace{0.2cm}}.

By extrapolating these interpretations to the analysis in \hyperref[e18c]{E.18c} of \mapu{pifaleymün}, all possibilities for the valency of the verb \mapu{pi-} are conceivable. That is, it can be transitive, intransitive, or ambitransitive, presenting two senses, one transitive and one intransitive. This phenomenon is also observed in the verb \mapu{nge-}, which is examined in greater detail in \S\hyperlink{sec.03.4.2.3}{3.4.2.3}.

Now, considering \hyperref[e19b]{E.19b}, it is evident that the analysis involving the co-occurrence of \textsc{DP+CA} (demonstrative + causative), equivalent to \mapu{-fa-l}, presents some challenges. In \mapu{Mapudüngun}, the mobility of suffixes\footnote{The mobility of suffixes refers to the relative positioning of the morpheme within the verbal form. This position is determined in relation to other suffixes and their proximity or distance from the root. Suffixes indicating mood, number, and person always appear after all other affixes that have been applied to the verb, meaning they are positioned furthest from the root, and their order among themselves does not change. However, with other semantically or grammatically related suffixes, the order can be interchangeable, as some suffixes are quite volatile or mobile.\vspace{0.2cm}} and the incorporation of roots (both nominal and non-nominal), even after a few suffixes following the initial root of the verb form, is not uncommon. However, it is less likely that this is the case here due to the position the affix would need to move to, which is away from its primary place of realisation. Therefore, it seems more plausible to think that the conjunction of the demonstrative and the causative, \mapu{-fa-l}, has generated the (bi)morpheme \mapu{-fal}, as \citeauthor{smeets2008} (\citeyear{smeets2008}: 274) herself speculates: ``the suffix \mapu{fal-} may have derived from \mapu{fa-} `become like this' and contain the causative \mapu{-(ü)l} (\mapu{-fa-l} `cause to become like this')''. This hypothesis would imply the transitive valency of the verb \mapu{pi-}, while also dissociating the co-occurrence of the causative with transitive roots, at least in this instance. Moreover, as previously mentioned, the meaning of both the bimorphemic \mapu{-fa-l} and the syncretic \mapu{-fal} coincide. \citeauthor{smeets2008} (\citeyear{smeets2008}: 272) has noted that \mapu{-fal} can imply that the subject orders someone to perform the action. In the analysis presented, the causative applies to the demonstrative that accounts for the action of the verb and is interpreted as `cause what the verb indicates', for example, in \mapu{pifaleymün}, `say + this (\mapu{fa-}) + to cause (\mapu{-l})', translated as `send (somebody) to say (something to somebody else)'.

\paragraph*{3.4.2.3\quad\mapu{nge-} -IV `be', -TV `have'}\hypertarget{sec.03.4.2.3}{}

While verifying the combination of verbal roots with the causative suffix \mapu{-(ü)l} in the memoirs of the \mapu{longko Paskwal Koña} \citep{mosbach1930}, the valency behaviour of these roots was also examined. Specifically, it was determined whether the roots were used only transitively or intransitively, or if they were ambivalent in their usage.

The root \mapu{nge-} exhibits bivalence due to its capacity to bear distinct interpretations, namely `have' in its transitive sense and `be' in its intransitive sense when it can combine with the causative suffix (\hyperref[e20a]{E.20a}). It is noteworthy that this root employs suffixes that are characteristic of transitive verbs when displaying its transitive sense, in the same way for its intransitive sense.

\paragraph{\example{20} \label{e20}}~
\vspace{-10pt}
\begin{enumerate}[label=\alph*.]
\item \label{e20a} \mapuex{nge \hspace{10pt} \textbf{-l} \hspace{6pt} -me \hspace{2pt} -fi \hspace{8pt} -ñ} \hfill (\citealp{smeets2008}: 126; 28)\\
      \gloss{IV.be \textbf{+CA} +TH +3P +IND1SG}\\
      `I have taken them there' -- `las he llevado allá'; \textit{literal:} `I made them be there'
\item \label{e20b} \gloss{IV.be \textbf{+MIO} +TH +3P +IND1SG}
\end{enumerate}

In \hyperref[e20b]{E.20b}, an alternative analysis is offered where the form \mapu{-l} is interpreted as `More Involved Object' (\textsc{MIO}) rather than causative (\textsc{CA}). This interpretation is proposed because only this example is found in \cite{smeets2008} where there is insufficient evidence to support the causative analysis. Additionally, no examples combining \mapu{nge-} with \mapu{-l} were found in the \mapu{longko Paskwal Koña}'s memoirs \citep{mosbach1930}, making it unlikely that the anthropologist's proposed analysis is correct. \citeauthor{golluscio2007} (\citeyear{golluscio2007}: 216) categorises the verb \mapu{nge-} among prototypical existence verbs that `do not allow causativisation', though this restriction is confirmed only with the causative \mapu{-(ü)m}, without mentioning if it extends to the causative \mapu{-(ü)l}.

The analysis presented in \hyperref[e20b]{E.20b} is plausible and does not contradict the interpretation of the verbal form. According to \citeauthor{smeets2008} (\citeyear{smeets2008}: 287), the morpheme \textsc{MIO} (More Involved Object) indicates a more direct, intense, or complete involvement of the patient in the event denoted by the verb (or the situation). Thus, the interpretation that could be derived from this suffix is something along the lines of `I have taken them [specifically these and no others] there'.

\subsection{\label{sec.03.5} Labile Roots and Person Agreement Suffixes}

In seeking to determine the valency of base verbal forms through morphology —that is, based on the combinations of roots and suffixes, specifically those that increase valency— it has been observed that some intransitive roots are also used transitively with the addition of person agreement suffixes that mark either the direct or inverse transition of the action, without the need for other valency modifiers. This observation could confirm that the roots generating these forms are, in fact, labile\footnote{Regarding the morpheme \mapu{-fi}, Zúñiga suggests the following references: (1) \citealp{salas1992a}, Chap. VI: \mapu{-fi} and \mapu{-e} are described as markers of focal and satellite person (a kind of obviative person, as known in Algonquian languages and others); both are person markers (\mapu{-fi} for 3\textsuperscript{rd} person definite patient and \mapu{-e} for other persons). \cite{salas1992a} considers \mapu{-e...-(m)ew} as a discontinuous suffix for 3\textsuperscript{rd} person focal. (2) \citealp{deaugusta1903}: This study considers this verbal morphology in terms of ``transitions``. According to the author: ``Transitions refer to the changes the verb undergoes in its ending depending on the person from whom the action originates and to whom it is directed... The four particles used for this purpose are \mapu{u, mo (mu), fi, e}'' (p. 66). (3) \citealp{zuniga2006}, Chap. III, Sect. 2.2 and Chap. IV, Sect. 2.2; \citeyear{zuniga2010b}, \citeyear{zuniga2015}: These texts explicitly analyse \mapu{-fi} as a Differential Object Marker (3\textsuperscript{rd} Person marker) and \mapu{-e} as an Inverse Marker. They also clarify that these are inflectional rather than derivational suffixes.\vspace{0.2cm}} as they behave intransitively when not co-occurring with these morphemes, or transitively with them —to prove lability no other valency increasing suffixes should appear in the verb form—; see examples \hyperref[e16a]{E.16a} and \hyperref[e16b]{E.16b}; also see the following examples:

\paragraph{\example{21} \label{e21}}~
\vspace{-10pt}
\begin{enumerate}[label=\alph*.]
\item \label{e21a} \mapuex{aye \hspace{26pt} -ka \hspace{12pt} \textbf{-fi} \hspace{8pt} -ñ} \hfill (\citealp{smeets2008}: 298; 9)\\
      \gloss{IV.laugh +CONT \textbf{+3P} +IND1SG}\\
      `I joked with him' -- `bromeé con él'; \textit{literal:} `I was laughing it'
\item \label{e21b} \mapuex{kewa \hspace{18pt} \textbf{-e} \hspace{10pt} -y \hspace{8pt} -ø \hspace{2pt} -ew} \hfill (\citealp{smeets2008}: 156; 11)\\
      \gloss{IV.fight \textbf{+INV} +IND +3 +3A}\\
      `he beat him' -- `él lo golpeó'
\item \label{e21c} \mapuex{kewa \hspace{16pt} -ya \hspace{8pt} \textbf{-fi} \hspace{8pt} -ñ} \hfill \citep{mosbach1930}\\
      \gloss{IV.fight +FUT \textbf{+3P} +IND1SG}\\
      `lo derrotaré'; \textit{literal:} `I will combat him'
\item \label{e21d} \mapuex{llüka \hspace{16pt} -ya \hspace{8pt} \textbf{-e} \hspace{8pt} -y \hspace{10pt} -ø \hspace{4pt} -u \hspace{8pt} -ø} \hfill (\citealp{smeets2008}: 194; 57)\\
      \gloss{IV.fear +FUT \textbf{+INV} +IND +1 +DL +1t2A}\\
      `I'm afraid of you' -- `tengo miedo de ti'; \textit{literal:} `you will scare me / I will fear you'
\item \label{e21e} \mapuex{llüka \hspace{14pt} \textbf{-fi} \hspace{8pt} -ø\footnote{This case does not involve a null morpheme, as it might appear, but rather the fusion of the \mapu{-i} from the indicative with the \mapu{-i} from the Differential Object Marker \mapu{-fi}.\vspace{0.2cm}} \hspace{2pt} -ø} \hfill \citep{mosbach1930}\\
      \gloss{IV.fear \textbf{+3P} +IND +3}\\
      `I was afraid of him'; \textit{literal:} `I feared him/it' -- `temíalo'
\item \label{e21f} \mapuex{maychü \hspace{28pt} \textbf{-fi} \hspace{8pt} -ñ} \hfill (\citealp{smeets2008}: 288; 8)\\
      \gloss{IV.rise-hands \textbf{+3P} +IND1SG}\\
      `I waved at him' -- `Lo saludé'
\item \label{e21g} \mapuex{meke \hspace{32pt} -a \hspace{12pt} \textbf{-e} \hspace{8pt} -n \hspace{20pt} -ew} \hfill (\citealp{smeets2008}: 484; 6)\\
      \gloss{IV.get-busy +FUT \textbf{+INV} +IND1SG +3A}\\
      `I will be busy at/on it' -- `estaré ocupado en/con eso'; \textit{literal:} `I will take care of it' -- `*me le ocuparé' (see footnote \ref{altisonante})
\item \label{e21h} \mapuex{meke \hspace{30pt} \textbf{-fi} \hspace{8pt} -ø \hspace{8pt} -ø} \hfill \citep{mosbach1930}\\
      \gloss{IV.get-busy \textbf{+3P} +IND +3}\\
      `hacerlo'; \textit{literal:} `get busy on/with\footnote{Do what is being mentioned, be occupied with doing it.\vspace{0.2cm}}'
\item \label{e21i} \mapuex{yewe \hspace{44pt} -ke \hspace{12pt} \textbf{-fwi} \hspace{12pt} -n} \hfill (\citealp{smeets2008}: 431; 53)\\
      \gloss{IV.be-ashamed +HAB +RI\textbf{+3P} +IND1SG}\\
      `I was ashamed before them' -- `estaba avergonzado ante ellos'; \textit{literal:} `if they embarrass me\footnote{Although this translation is not semantically accurate, it is grammatically correct.\vspace{0.2cm}}'
\end{enumerate}

\subsubsection{\label{sec.03.5.1} Labile Roots Incorporated into \mapu{Düngupeyüm}}

Regarding the \mapu{Düngupeyüm} morphological analysis system, from now on, labile roots will be included in both lists of verbal roots, transitive and intransitive, with the appropriate sense to the root valency. This will respectively increase the ambiguity in the analysis of forms containing these roots. However, failing to include these roots in both lists would mean missing out on potential analyses. At the same time, the morphological analyser system has been adjusted so that the causatives, either of them, \mapu{-(ü)m} (\S\ref{sec.03.3}) or \mapu{-(ü)l} (\S\ref{sec.03.4}), do not co-occur with transitive roots. In other words, for labile roots, the causatives will only co-occur when these roots are expressing their intransitive sense. The following labile roots were collected in the system during the course of this study:

\begin{enumerate}[label=\arabic*.]
\item \mapu{aye-} `to laugh' (\textsc{IV})\\
      \mapu{aye-} `to laugh at; to laugh with' (\textsc{TV})
      \vspace{0.1em}
\item \mapu{kewa-} `to fight, to quarrel, to struggle, to combat' (\textsc{IV})\\
      \mapu{kewa-} `to hit, to strike, to defeat, to overcome' (\textsc{TV})
      \vspace{0.1em}
\item \mapu{llüka-} `to get frightened, to be scared, to fear, to be afraid' (\textsc{IV})\\
      \mapu{llüka-} `to scare, to frighten' (\textsc{TV})
      \vspace{0.1em}
\item \mapu{meke-} `to take care of, to deal with, to be busy, to last, to endure' (\textsc{IV})\\
      \mapu{meke-} `to make' (\textsc{TV})
      \vspace{0.1em}
\item \mapu{monge-} `to get life, to heal, to recover' (\textsc{IV})\\
      \mapu{monge-} `to revive, to bring back to life, to heal' (\textsc{TV})
      \vspace{0.1em}
\item \mapu{nge-} `to be' (\textsc{IV})\\
      \mapu{nge-} `to have' (\textsc{TV}) (see \S\hyperlink{sec.03.4.2.3}{3.4.2.3})
      \vspace{0.1em}
\item \mapu{püna-} `to stick, to glue, to adhere' (\textsc{IV})\\
      \mapu{püna-} `to glue, to paste' (\textsc{TV})
      \vspace{0.1em}
\item \mapu{waychüf-} `to fall, to get turned over, to roll, to spin' (\textsc{IV})\\
      \mapu{waychüf-} `to discard, to flip, to roll, to spin' (\textsc{TV})
      \vspace{0.1em}
\item \mapu{yewe-} `to feel embarrassed, to be ashamed' (\textsc{IV})\\
      \mapu{yewe-} `to respect' (\textsc{TV})
\end{enumerate}

\subsection{\label{sec.03.6} Labile Roots and Other Suffixes}

One observation that has drawn our attention is that some labile roots, when presenting their transitive sense, always or very frequently appear with a suffix that, in principle, does not affect the valency of the verbal root. For instance, \mapu{kewa-}, whose intransitive meaning is `fight, struggle, combat' (\hyperref[e22a]{E.22a} and \hyperref[e22b]{E.22b}), occurs in its transitive sense, `hit, strike', with the suffix indicating `Habitual Aspect', \mapu{-ke} (\hyperref[e22c]{E.22c} and \hyperref[e22d]{E.22d}):

\paragraph{\example{22} \label{e22}}~
\vspace{-10pt}
\begin{enumerate}[label=\alph*.]
\item \label{e22a} \mapuex{epe \hspace{54pt} kewa \hspace{26pt} -fu \hspace{4pt} -ø \hspace{8pt} -ø \hspace{4pt} -iñ}\\
      \gloss{AV.almost IV.fight\_combat +RI +IND +1 +PL}\\
      `we almost would have fought' -- `casi hubiéramos peleado'
\item \label{e22b} \mapuex{ka \hspace{68pt} kewa \hspace{30pt} -i \hspace{8pt} -ng \hspace{2pt} -ün}\\
      \gloss{AJ.(an)other IV.fight\_combat +IND +3 +PL}\\
      `they fought again' -- `pelearon otra vez'
\item \label{e22c} \mapuex{kewa \hspace{54pt} \textbf{-ke} \hspace{6pt} -fu \hspace{4pt} -y \hspace{10pt} -ø \hspace{6pt} ñi \hspace{12pt} kure}\\
      \gloss{IV.fight\_combat \textbf{+HAB} +RI +IND +3 SP.his NN.wife}\\
      `[he used to] hit his wife' -- `[le] pegaba a su mujer'
\item \label{e22d} \mapuex{welu \hspace{112pt} kewa \hspace{26pt} -nge \hspace{6pt} \textbf{-ke} \hspace{8pt} -la \hspace{10pt} -y \hspace{10pt} -ø}\\
      \gloss{CJ.on-the-contrary\_but IV.fight\_combat +PASS \textbf{+HAB} +NEG +IND +3}\\
      `but they don't usually hit her' -- `pero no suelen pegarle';\\
      \textit{literal:} `but [she] is not constantly hit' -- `pero no es constantemente golpeada'
\end{enumerate}

The habitual aspect morpheme \mapu{-ke} signifies the ongoing nature of the action expressed by the verb. Depending on the entity to which it is applied, it can denote a characteristic or a habit of the referent. While researchers of the \mapu{Mapuche} language have not focused extensively on this suffix, they agree that a verb containing it takes on an imperfective form that indicates habituality, and that this affix is quite common (\citealp{zuniga2006}: 160).

Another root where this co-occurrence of the transitive sense of the verbal root and the suffix \mapu{-ke} is observed is \mapu{kon-} `enter', whose corresponding transitive sense is `begin, start' (\hyperref[e23]{E.23}).

\paragraph{\example{23} \label{e23}}~
\vspace{-10pt}
\begin{enumerate}[label=\alph*.]
\item[] \mapuex{kon \hspace{22pt} \textbf{-ke} \hspace{8pt} -fu \hspace{4pt} -y \hspace{6pt} -ng \hspace{1pt} -ün \hspace{6pt} ñi \hspace{16pt} kekaw \hspace{18pt} -a \hspace{12pt} -l} \hfill (\citealp{smeets2008}: 203; 121)\\
\gloss{IV.enter \textbf{+HAB} +RI +IND +3 +PL SP.his IV.complain +FUT +OVN}\\
`they usually started to complain' -- `ellos generalmente comenzaban a quejarse'; \textit{literal:} `they began their complaint'
\end{enumerate}

The root \mapu{kon-} `enter' is not generally considered labile, though it can exhibit a transitive sense in specific contexts, as seen in \hyperref[e23]{E.23}. Here, the verb's complement is an inanimate object formed by a nominalised or non-finite verb\footnote{This is another clarification for which we owe thanks to Fernando Zúñiga and his deep knowledge of \mapu{Mapudüngun}.\vspace{0.2cm}}. Often, the transitive interpretation in translations aligns the \mapu{Mapudüngun} expression with Spanish idiomatic usage ('they started...'); but \hyperref[e23]{E.23} could be more literally translated as `they entered into their complaint', which highlights its intransitive sense. Additionally, many instances of \mapu{kon-} with the marker \mapu{-ke} do not allow a transitive interpretation unless combined with suffixes typically used for transitive forms (\hyperref[e24]{E.24}). However, when a transitive sense is possible, the morpheme \mapu{-ke} is present.

\paragraph{\example{24} \label{e24}}~
\vspace{-10pt}
\begin{enumerate}[label=\alph*.]
\item[] \mapuex{kon \hspace{22pt} \textbf{-ke} \hspace{8pt} -fu \hspace{6pt} -y \hspace{8pt} -ø \hspace{6pt} epu \hspace{8pt} wentru} \hfill \citep{mosbach1930}\\
\gloss{IV.enter \textbf{+HAB} +RI +IND +3 NU.two NN.man}\\
`two men would enter'; \textit{literal:} `enter used to two men'
\end{enumerate}

\subsection{\label{sec.03.7} Transitiviser Suffix \mapu{-tu}}

The suffix \mapu{-tu} is often assumed to transitivise intransitive roots or complex verbal themes, among other functions (§\ref{sec.03.7.1}). However, \citeauthor{smeets2008} (\citeyear{smeets2008}: 297) states that \mapu{-tu} can be added to both intransitive and transitive roots. With intransitive verbs, \mapu{-tu} adds an object, resulting in a transitive form. With transitives, it incorporates an additional object, creating a form with two objects. Despite its name, \mapu{-tu} does not always transitivise\footnote{Transitive verbs select objects; intransitive verbs do not (\citealp{tubino2017}: 20).\vspace{0.2cm}}; with transitive roots, it incorporates an object rather than altering valency. This function is not unique to \mapu{-tu}, as causatives (§\ref{sec.03.3}, §\ref{sec.03.4}), applicatives\footnote{Applicatives modify verb forms to include additional participants (\citealp{zuniga2006}: 123).\vspace{0.2cm}} (§\ref{sec.03.7.1}), and benefactives also increase valency.

\paragraph{\example{25} \label{e25}}~
\vspace{-10pt}
\begin{enumerate}[label=\alph*.]
\item \label{e25a} \mapuex{anel-} `(Vt) threaten, menace' -- `amenazar, intimidar' \hfill (\citealp{smeets2008}: 496)\\
      \mapuex{anel-\textbf{tu}} `(Vt) threaten someone with something' -- `amenazar a alguien con algo'
\item \label{e25b} \mapuex{anel \hspace{36pt} \textbf{-tu} \hspace{6pt} -la \hspace{8pt} -e \hspace{10pt} -y \hspace{10pt} -ø \hspace{4pt} -u \hspace{8pt} -ø} \hfill (\citealp{smeets2008}: 157; 21)\\
      \gloss{TV.threaten \textbf{+TR} +NEG +INV +IND +1 +DL +1t2A}\\
      `I did not threaten you' -- `no te amenacé'
\end{enumerate}

In the verbal form represented in \hyperref[e25b]{E.25b}, there are only two arguments: agent and patient (direct object). There is no other (indirect) object explicitly present, as would be expected according to the explanation in the \hyperref[{sec.03.7}]{paragraph above} and the definition of \mapu{anel-\textbf{tu}} provided in \hyperref[e25a]{E.25a}. It is possible that the example is incomplete and that the verbal form of the sentence was only used to explain the `dative subject' (agent) morpheme occurring within it. Therefore, it is necessary to examine another case:

\paragraph{\example{26} \label{e26}}~
\vspace{-10pt}
\begin{enumerate}[label=\alph*.]
\item \label{e26a} \mapuex{are-} `(Vt) lend to' -- `prestar a';\hfill (\citealp{smeets2008}: 498)\\
\mapuex{are-l} `lend to' -- `prestar a' it is more frequent; \mapuex{arelenew kiñe mansun} `he lent me one ox' -- `me prestó un buey';\\
      \mapuex{are-\textbf{tu}} `(Vt) borrow from' – `pedir prestado a';\\
      \mapuex{aretuenew ketran} $\sim$ \mapuex{aretuketranenew} `he borrowed wheat from me' -- `me pidió prestado trigo'
\item \label{e26b} \mapuex{are \hspace{20pt} \textbf{-tu} \hspace{2pt} -n \hspace{14pt} -mew \hspace{2pt} monge \hspace{2pt} -li \hspace{6pt} -y \hspace{8pt} -ø} \hfill (\citealp{smeets2008}: 62; 7)\\
      \gloss{TV.lend \textbf{+TR} +PVN +INST IV.live +ST +IND +3}\\
      `he lives of borrowing' -- `él vive de pedir prestado'; \textit{literal:} `he's living off the borrowed' -- `de lo prestado está viviendo'
\item \label{e26c} \mapuex{fey \hspace{30pt} re \hspace{22pt} are \hspace{8pt} \textbf{-tu} \hspace{4pt} -n \hspace{24pt} ropa \hspace{24pt} nie \hspace{14pt} -y \hspace{8pt} -ø} \hfill (\citealp{smeets2008}: 198; 90)\\
      \gloss{DP.that AV.only TV.lend \textbf{+TR} +PVN NN.clothing TV.have +IND +3}\\
      `he has only borrowed clothes' -- `él solamente ropa prestada tiene'
\end{enumerate}

In \hyperref[e26]{E.26}, both forms, \mapu{are-} `lend' and \mapu{are-tu} `borrow,' imply ``someone VERB something to/from someone''; however, in each example, \hyperref[e26b]{E.26b} and \hyperref[e26c]{E.26c}, the expected objects are retained, and no additional object is introduced.

Returning to the description of the suffix functions, despite the name chosen to identify the form \mapu{-tu}, one might think that a base verb occurring with \mapu{-tu} and presenting a single object is formed from an originally intransitive verbal root, with \mapu{-tu} adding that object and thereby transforming it into a transitive verb. However, as demonstrated by \hyperref[e25]{E.25} and \hyperref[e26]{E.26}, the addition of an object does not always occur.

\subsubsection{\label{sec.03.7.1} Applicative Suffix \mapu{-tu}}

Moreover, there are additional aspects to consider for the suffix \mapu{-tu}. For instance, \citeauthor{salas1992a} (\citeyear{salas1992a}: 150) notes that it is a `repetitive-inversive' suffix (\hyperref[e27a]{E.27a}), although at times it merely indicates `inversion' (\hyperref[e27b]{E.27b}) or `repetition' (\hyperref[e27c]{E.27c}), and occasionally suggests that the action originates as a consequence of a previous one (\hyperref[e27d]{E.27d}). He also explains that it functions as a non-causative transitiviser (\hyperref[e27e]{E.27e}, \hyperref[e27f]{E.27f}) (ibid: 187) or is added to verbal stems formed by reduplicated roots\footnote{The specific function of the suffix in this context is not detailed in the cited text.\vspace{0.2cm}} (ibid: 189) (\hyperref[e27g]{E.27g}). Additionally, it attaches to verbal stems formed from nominal roots, where it acts as a verbaliser, adding a somewhat non-specific semantic nuance but conveying the sense of `performing the action appropriate to such a noun' (\hyperref[e27h]{E.27h} to \hyperref[e27k]{E.27k}). \cite{smeets2008} reports similar functions for this suffix, though she places it differently across the slots occupied by verb-forming suffixes, suggesting it could either be homophonous suffixes or a single suffix fulfilling various functions depending on its position.

\paragraph{\example{27} \label{e27}} \hfill {\scriptsize (All these examples have been drawn from \citealp{salas1992a}. Only the source page is indicated for each example.)}
\vspace{-10pt}
\begin{enumerate}[label=\alph*.]
\item \label{e27a} \mapuex{küpa-y} `he came' \hfill (p. 150)\\
      \mapuex{küpa\textbf{-tu}-y} `he came (back)'
\item \label{e27b} \mapuex{kansa}\footnote{From the Spanish `\textit{cansar(se)}'.\vspace{0.2cm}}\mapu{-n} `get tired' \hfill (p. 150)\\
      \mapuex{kansa\textbf{-tu}-n} `I rested'
\item \label{e27c} \mapuex{wew-nge-y-m-i} `you were beaten' \hfill (p. 150)\\
      \mapuex{wew-nge\textbf{-tu}-y-m-i} `you were beaten again'
\item \label{e27d} \mapuex{wikürü-y} `it tore' \hfill (p. 150)\\
      \mapuex{wikürü\textbf{-tu}-y} `It tore (after that, as a consequence of that)'
\item \label{e27e} \mapuex{tripa-n} `go out' \hfill (p. 187)\\
      \mapuex{tripa\textbf{-tu}-n} `intercept (someone)' -- `salirle al paso (a alguien)'
\item \label{e27f} \mapuex{rüngkü-n} `jump' \hfill (p. 187)\\
      \mapuex{rüngkü\textbf{-tu}-n} `pounce on (someone)' -- `saltarle (a alguien) encima'
\item \label{e27g} \mapuex{rüngkü-rüngkü\textbf{-tu}-y} `he bounced' -- `dio saltos' \hfill (p. 189)
\item \label{e27h} \mapuex{kofke} `bread' \hfill (p. 194)\\
      \mapuex{kofke\textbf{-tu}-n} `eat bread'
\item \label{e27i} \mapuex{koyla} `lie (noun)' \hfill (p. 194)\\
      \mapuex{koyla\textbf{-tu}-n} `lie (verb)'
\item \label{e27j} \mapuex{kawello} `horse' \hfill (p. 194)\\
      \mapuex{kawello\textbf{-tu}-n} `ride (on horseback)'
\item \label{e27k} \mapuex{kütral} `fire' \hfill (p. 194)\\
      \mapuex{kütral\textbf{-tu}-n} `set fire, light a fire'
\end{enumerate}

\cite{golluscio2010} and \citeauthor{zuniga2009b} (\citeyear{zuniga2009a}, \citeyear{zuniga2009b}, \citeyear{zuniga2011}, \citeyear{zuniga2024}) describe the suffix \mapu{-tu} as `Applicative', noting its multiple functions. While a detailed discussion is beyond this article's scope, it is worth summarising that \mapu{-tu} can increase, have a neutral effect on, or decrease valency, depending on various factors. Thus, examining forms with the morpheme \mapu{-tu} to determine the valency of base verbs is not particularly effective.

To conclude the exploration of the valency behaviour of verbs where the suffix \mapu{-tu} occurs, it is important to note the multiple functions of this suffix, a point on which the cited linguists agree. In \citeauthor{zuniga2009b}'s words (\citeyear{zuniga2009b}: 10):

\begin{quote} \label{qt11zuniga}
Note that \mapu{-tu} exhibits a broad range of possible meanings as a non-applicative formative: it functions as a denominal verbaliser (e.g., \mapu{mamüll-tu} `gather firewood', from \mapu{mamüll} `firewood'), but it is also used to form the iterative (e.g., \mapu{rüngkü-y} `he/she jumped' vs. \mapu{rüngkü-rüngkü-tu-y} or \mapu{rüngkü-rüngkü-nge-y} `he/she bounced, jumped repeatedly'), and frequently simply means `return' (e.g., \mapu{aku-y} `he/she arrived here' vs. \mapu{aku-tu-y} `he/she returned here'; note also the pair \mapu{amu-y} `he/she went' vs. \mapu{amu-tu-y} `he/she left/returned').
\end{quote}

It may be that the suffixes in question are not the same\footnote{From this paragraph until the end of section \S\ref{sec.03.7.1}, the discussion shifts away from the main focus of this article to address the functional and semantic aspects of the morpheme(s) \mapu{-tu}. These aspects would require a dedicated study from these perspectives.\vspace{0.2cm}}, despite sharing a common origin. For instance, one suffix might have specialised in certain meanings when close to the root, while in its position towards the end of the verbal form, near the inflectional suffixes, it could have taken on an iterative/restorative meaning. This specialisation might have led, through its evolution, to the development of distinct suffixes. Alternatively, the suffix \mapu{-tu} in its more distal position, further from the root, might have a different origin\footnote{\citeauthor{zuniga2009b} (\citeyear{zuniga2009b}: 7) relates the affix form \mapu{-tu} to the verb \mapu{tu-} `take'.\vspace{0.2cm}}, possibly deriving from \mapu{tuw-} `come from, originate in, begin', which would lend it a meaning such as `originate again, start anew', or a similar interpretation. A couple of examples in \cite{smeets2008} might also suggest a different origin for the pair of \mapu{-tu} suffixes.

\paragraph{\example{28} \label{e28}}~
\vspace{-10pt}
\begin{enumerate}[label=\alph*.]
\item \label{e28a} \mapuex{pelo\footnote{For the etymology of \mapu{pelo} and similar lexicalised forms, see footnote 20 in \citealp{chandia2025}. For the classification of \mapu{pelo} as a lexicalised nominalised verb, see footnote f in Table 4 of the same work.\vspace{0.2cm}} \hspace{24pt} \textbf{-tu \hspace{2pt} -tu} \hspace{4pt} -y \hspace{8pt} -ø \hspace{2pt} -iñ} \hfill (\citealp{smeets2008}: 476; 62)\\
      \gloss{TV.light.up \textbf{+TR +RE} +IND +1 +PL}\\
      `we have seen the light again' -- `hemos vuelto a ver la luz'; \textit{literal:} `it lit up for us again'
\item \label{e28b} \mapuex{ina \hspace{24pt} \textbf{-tu \hspace{2pt} -tu} \hspace{2pt} -fi \hspace{6pt} -y \hspace{10pt} -ø \hspace{4pt} -iñ} \hfill (\citealp{smeets2008}: 476; 63)\\
      \gloss{AV.close \textbf{+TR +RE} +3P +IND +1 +PL}\\
      `we have also been close to our own [matter]' -- `también hemos estado cerca de lo nuestro';\\
      \textit{literal:} `we approached again to it'
\end{enumerate}

In the two previous examples, the homophonous suffixes \textsc{TR} (transitiviser) and \textsc{RE} (repetitive/restorative) co-occur according to the analysis provided by the author, suggesting the presence of two distinct suffixes. However, this source only provides these two examples. There are also two other instances where the morpheme \mapu{-tu} appears twice within the same verb form, but they were analysed differently:

\paragraph{\example{29} \label{e29}}~
\vspace{-10pt}
\begin{enumerate}[label=\alph*.]
\item \label{e29a} \mapuex{wiño \hspace{22pt} \textbf{-tu \hspace{2pt} -tu} \hspace{2pt} -fu \hspace{4pt} -y \hspace{8pt} -ø} \hfill (\citealp{smeets2008}: 410; 43)\\
      \gloss{IV.return \textbf{+RE +RE} +RI +IND +3}\\
      `they kept coming back' -- `seguían volviendo'
\item \label{e29b} \mapuex{kona -kona \hspace{34pt} \textbf{-tu \hspace{2pt} -tu} \hspace{4pt} -y \hspace{8pt} -m \hspace{2pt} -ün ?} \hfill (\citealp{smeets2008}: 468; 27)\\
      \gloss{NN.warrior -RNNR \textbf{+SFR +RE} +IND +2 +PL}\\
      `did you get courage again?' -- `tuvieron coraje nuevamente?'
\end{enumerate}

In the analysis presented for \hyperref[e29a]{E.29a}, \citeauthor{smeets2008} (\citeyear{smeets2008}: 410) interprets the reduplication of the Repetitive/Restorative suffix as a plausible explanation, though other interpretations are also possible. The same could apply to \hyperref[e29b]{E.29b}, where the initial occurrence of \mapu{-tu} is interpreted as a verbal theme marker for reduplicated roots, in this case, nominal ones (\textsc{RNNR}). According to \cite{smeets2008}'s analysis, the suffix \mapu{-tu} in this verb might initially appear to lack semantic content. However, it may actually share the iterative sense with the reduplicated root, thus reinforcing the notion of repetition. \citeauthor{zuniga2024} (\citeyear{zuniga2024}: 206) seems to concur with this interpretation, at least as presented in this paragraph.

\begin{quote} \label{qt12zuniga}
The non-applicative marker \mapu{-tu} is common and can, depending on the lexical base, function to antipassivise, telicise, or form a reversive/repetitive action; the latter is possibly related to one of the iterative formations of the root. In some verbs, the suffix \mapu{-tu} indicates increased intensity or greater impact on the patient.
\end{quote}

Finally, regarding the denominal verbaliser function of the suffix \mapu{-tu}, it seems that the actions it forms involve a certain duration, rather than occurring instantly, they occur in repeated instances. For example, \mapu{kofke} `bread' -- \mapu{kofke-tu} `eat bread': bread is eaten bite by bite; or \mapu{pulku} `wine' -- \mapu{pulku-tu} `drink wine': wine is drunk sip by sip, etc. (see \hyperref[e27]{E.27}). Therefore, it would not be a semantically empty suffix, and the meaning it conveys is also related to the repetitive/restorative aspect.

\section{\label{sec.04} Transitivising Properties of Various Suffixes}

Apart from the causatives \mapu{-(ü)m} (\S\ref{sec.03.3}) and \mapu{-(ü)l} (\S\ref{sec.03.4}), and the applicative \mapu{-tu} (\S\ref{sec.03.7}), other suffixes are reported to have transitivising properties in \mapu{Mapudüngun} verb forms. The researchers cited in this work also include the suffixes \mapu{-ma $\sim$ -(ü)ñma} and \mapu{-ye} among those that can transitivise intransitive verbal roots or increase the valency of transitive ones, similar to the suffix \mapu{-(l)el}\footnote{For details regarding this form, see the last paragraph of \S\ref{sec.03.4}.\vspace{0.2cm}}. This is illustrated in the following examples:

\paragraph{\example{30} \label{e30}}~
\vspace{-10pt}
\begin{enumerate}[label=\alph*.]
\item \label{e30a} \mapuex{lladkü-} `become sad' \hfill (\citealp{zuniga2009a}: 10)\\
      \mapu{lladkü\textbf{-ñma}} `offer condolences'
\item \label{e30b} \mapuex{ngüma-} `cry' \hfill (\citealp{zuniga2009a}: 3)\\
      \mapu{ngüma\textbf{-ye}} `mourn, grieve for, cry for [someone]'
\item \label{e30c} \mapuex{ngilla-} `buy' \hfill (\citealp{zuniga2009a}: 7)\\
      \mapu{ngilla\textbf{-lel}} `buy [something] on behalf of [somebody else]'
\end{enumerate}

\subsection{\label{sec.04.1} Transitivity Induced by the Aspectual Suffixes \mapu{-nie} and \mapu{-künu}}

In \mapu{Mapudüngun}, several suffixes can confer a transitive sense onto intransitive roots. For instance, aspectual modifiers such as \mapu{-nie} and \mapu{-künu}, identified by \citeauthor{smeets2008} (\citeyear{smeets2008}: 293) as `Progressive Persistent' and `Perfective Persistent', respectively, occupy the same slot, according to the author's description, thereby implying mutual exclusivity. Section \S\ref{sec.03.1} outlines the durative effect of these suffixes on verbs. \citeauthor{smeets2008} (\citeyear{smeets2008}: 294) also notes that these suffixes introduce meaning variations depending on whether they combine with telic or atelic verbs, and that verbs with these modifiers are transitive.

The relationship between the verb \mapu{nie-} `have' and the morpheme \mapu{-nie} is evident (\hyperref[e3]{E.3}, p. \pageref{e3}) \mapu{ngülüm\textbf{nie}y plata} `he saves money' could be paraphrased as `he \textbf{has} money accumulating', as evident is the connection between the verb \mapu{künu-} `leave' and the suffix \mapu{-künu} (\hyperref[e19b]{E.19b}, p. \pageref{e19b}) \mapu{ellka\textbf{künu}lelfalyengey} `they should \textbf{be left} hidden'. Note that both verbs are transitive. For this reason, \citeauthor{zuniga2009b} (\citeyear{zuniga2009b}: 8) proposes the plausible hypothesis that some lexical items have produced grammaticalised variants without losing their function and value as base verbs\footnote{\cite{fernandez-malvestitti2012} and \cite{fernandez2015} regard these forms as serial verb constructions, assigning verb status to the second component rather than treating it as a suffix. Their research focuses on certain variants of \mapu{Mapudüngun} spoken in Argentina, so there may be a perceptual difference between academic approaches to the \mapu{Mapuche} language or the elements may be at different stages of evolution in the various dialects of this language.\vspace{0.2cm}}. Indeed, it is often challenging to determine whether the verbal theme in forms containing these suffixes (and others with similar conditions) consists of verbal compounds where the second element, also a verb, modulates the action of the main verb\footnote{\citeauthor{smeets2008} (\citeyear{smeets2008}: 325) mentions that in compounds, posture verbs appear without the suffixes \mapu{-künu}, \mapu{-nie} or \mapu{-(kü)le}, giving the example \mapu{külü-nag-} `lean on the elbow'; \textit{literal:} `lean, tilt-down'. This may indicate that the verbs \mapu{künu-} `let' and \mapu{nie-} `have' are forming verbal compounds with the base verb, thereby modifying it, rather than functioning as aspectual suffixes.\vspace{0.2cm}}, or, from another perspective, if the verbal theme is formed solely by a verbal root that is subsequently modulated by the aspectual suffix. In the examples provided in \hyperref[e31]{E.31}, the section labelled `C' (from compound) illustrates how each form would appear by replacing the suffix \mapu{-nie} with the verb \mapu{nie-} `have', and the suffix \mapu{-künu} with the verb \mapu{künu-} `let'.

Regarding transitivisation, and in line with \citeauthor{smeets2008} (\citeyear{smeets2008}: 294) observations, the following examples are presented:

\newpage
\paragraph{\example{31} \label{e31}}~
\vspace{-10pt}
\begin{enumerate}[label=\alph*.]
\item \label{e31a} \mapuex{angid \hspace{16pt} \textbf{-nie} \hspace{8pt} -fi \hspace{8pt} -n} \hfill (\citealp{smeets2008}: 294; 5)\\
      \gloss{IV.blaze \textbf{+PRPS} +3P +IND1SG}\\
      `I keep it scorching' -- `Lo mantengo ardiendo'; \textit{literal:} `I have it blazing'\\
      \small{(C: \gloss{IV.blaze \textbf{+TV.have} +3P +IND1SG} → `I have it blazing')}
\item \label{e31b} \mapuex{aye \hspace{24pt} \textbf{-nie} \hspace{10pt} -a \hspace{18pt} -fe \hspace{14pt} -y \hspace{10pt} -ø \hspace{4pt} -u \hspace{8pt} -ø} \hfill (\citealp{smeets2008}: 252; 7)\\
      \gloss{IV.laugh \textbf{+PRPS} +FUT +RI+INV +IND +1 +DL +1t2A}\\
      `I laughed at you' -- `me reí de ti': \textit{literal:} `I would have laugh/laughed [at] you'\\
      \small{(C: \gloss{IV.laugh \textbf{+TV.have} +FUT +RI+INV +IND +1 +DL +1t2A} → `I would have laugh/laughed [at] you')}
\item \label{e31c} \mapuex{weyel \hspace{12pt} \textbf{-künu} \hspace{2pt} -fi \hspace{8pt} -n} \hfill (\citealp{smeets2008}: 168; 57)\\
      \gloss{IV.swim \textbf{+PFPS} +3P +IND1SG}\\
      `I let him swim' -- `lo dejé nadar'\\
      \small{(C: \gloss{IV.swim \textbf{+TV.let} +3P +IND1SG} → `I let him swim')}
\item \label{e31d} \mapuex{weyel \hspace{12pt} \textbf{-nie} \hspace{6pt} -fi \hspace{8pt} -n} \hfill (\citealp{smeets2008}: 168; 56)\\
      \gloss{IV.swim \textbf{+PRPS} +3P +IND1SG}\\
      `I kept him swimming' -- `lo tuve/mantuve nadando'\\
      \small{(C: \gloss{IV.swim \textbf{+TV.have} +3P +IND1SG} → `I had/kept him swimming')}
\item \label{e31e} \mapuex{wirar \hspace{18pt} \textbf{-nie} \hspace{8pt} -e \hspace{10pt} -n \hspace{22pt} -ew} \hfill (\citealp{smeets2008}: 295; 7)\\
      \gloss{IV.shout \textbf{+PRPS} +INV +IND1SG +3A}\\
      `he is shouting at me' -- `él me está gritando'; \textit{literal:} `it has/keeps [me] shouting at me'\\
    \small{(c: \gloss{IV.shout \textbf{+TV.have} +INV +IND1SG +3A} → `it has/keeps [me] shouting at me')}
\end{enumerate}

Nevertheless, of all the intransitive verbs that were found interacting with one of these suffixes in the consulted works (see \S\ref{sec.02.1}), none produce a transitive form in the complete verbal construction in which they participate. This is likely because these forms also include the reflexive suffix \mapu{-(u)w}, which reduces the valency. Indeed, \cite{smeets2008} herself classifies these verbal themes as intransitive —no intransitive verbal roots (base verbs) were found combining solely with the applicative \mapu{-nie}; the reflexive \mapu{-(u)w} is always present in the examples of her work:

\paragraph{\example{32} \label{e32}} \hfill {\scriptsize (From \cite{smeets2008}'s dictionary. pp. 489--580)}
\vspace{-10pt}
\begin{enumerate}[label=\alph*.]
\item \label{e32a} \mapuex{külü-künu-w} `(Vi) lie to one side' -- `recostarse de lado'
\item \label{e32b} \mapuex{llikosh-künu-w} `(Vi) squat down, crouch down' -- `ponerse en cuclillas, agacharse'
\item \label{e32c} \mapuex{payla-künu-w} `(Vi) lie down on one's back' -- `recostarse de espaldas'
\item \label{e32d} \mapuex{reka-künu-w} `(Vi) adopt a position with the legs apart' -- `ponerse con las piernas separadas'
\item \label{e32e} \mapuex{rekül}\footnote{The verb \mapu{rekül-} could be a lexical loan; it indeed resembles the Catalan verb \textit{recolzar-se} (lean or support oneself), both in form and meaning.\vspace{0.2cm}}\mapu{-künu-w} `(Vi) sit/lay down leaning (upon one's elbow/on one's arm/against a wall)' -- `sentarse o recostarse apoyándose (sobre el codo, sobre el brazo o contra la pared)'
\item \label{e32f} \mapuex{trana-künu-w} `(Vi) fall on the floor' -- `caer(se) al suelo'
\item \label{e32g} \mapuex{treka-künu-w} `(Vi) set off, set out for' -- `partir, echarse a andar'
\item \label{e32h} \mapuex{ünif-künu-w} `(Vi) lie down all stretched out' -- `recostarse totalmente estirado'
\item \label{e32i} \mapuex{wille/wüllü-künu-} `(Vi) urinate' -- `orinar\footnote{It is possible that \mapu{wüllü-künu-} means `urinate (on oneself)', considering that the form \mapu{künu} conveys the sense of `let', and that in the CORLEXIM \citep{chandia2014} dictionaries \mapu{wüllü-tu} and \mapu{wüllü-n} are reported as meaning `urinate'.\vspace{0.2cm}}'
\item \label{e32j} \mapuex{wira-künu-w} `(Vi) adopt a position with the legs apart' -- `ponerse con las piernas separadas'
\item \label{e32k} \mapuex{witra-künu-w} `(Vi) get up and stay put, come to a stand still'; \mapu{witrakünuwnge!} `get up and stand still!'; \mapuex{iñche trekalefun, perumen fentren che, fey witrakünuwün} `[while] I was walking, I suddenly saw a lot of people [and] then I stopped' -- `levantarse y quedarse quieto, quedarse parado; \mapuex{witrakünuwnge!} `levántate y quédate quieto!'; \mapuex{iñche trekalefun, perumen fentren che, fey witrakünuwün} `[mientras] caminaba, de pronto vi mucha gente [y] entonces paré' -- \textit{literal:} `yo caminaba rápido, repentinamente vi mucha gente, entonces me quedé quieto'
\end{enumerate}

Only two instances were found of verb forms generated from an intransitive root and exclusively transitivised by one of the applicatives \mapu{-nie} or \mapu{-künu}; however, as previously mentioned, it is also possible that these are verbal compounds (verbal themes formed by two verb roots):

\paragraph{\example{33} \label{e33}}~
\vspace{-10pt}
\begin{enumerate}[label=\alph*.]
\item \label{e33a} \mapuex{maychi \hspace{30pt} \textbf{-nie} \hspace{8pt} -y \hspace{8pt} -ø \hspace{8pt} ñi \hspace{12pt} kug} \hfill \citep{mosbach1930}\\
      \gloss{IV.rise-hands \textbf{+PRPS} +IND +3 SP.his NN.hand}\\
      `he extended his hands'; \textit{literal:} `he had his hand raised'
\item \label{e33b} \mapuex{trongo \hspace{14pt} \textbf{-künu} \hspace{4pt} -y \hspace{6pt} -ø \hspace{20pt} tol} \hfill (\citealp{smeets2008}: 296; 18)\\
      \gloss{IV.frown \textbf{+PFPS} +IND +3 NN.forehead}\\
      `he frowned, thereby causing wrinkles to appear on his forehead' --\\
      `él frunció el ceño, provocando la aparición de arrugas en su frente'
\end{enumerate}

Besides the possibility mentioned in the previous paragraph, a few points need to be addressed regarding these last two examples. In both instances, an object that is already implicit in the base verbal form is presented explicitly; in the first example, `hands' and `raise hands', and in the second, `forehead' and `frown'.

In \hyperref[e33a]{E.33a}, it is likely that \mapu{maychi-} is a compound formed by \mapu{may} `yes, assent' and the element \mapu{chi}, which in this context could be interpreted as a determiner `the assent' or an adjectiviser `assenting'. CORLEXIM \citep{chandia2014} lists \mapu{maychi-} or \mapu{maychü-} as `signal with hands'. Regarding \mapu{trongo} from \hyperref[e33b]{E.33b}, it does not appear in \cite{smeets2008}'s dictionary, nor was a definition found for this word in CORLEXIM \citep{chandia2014}. In a couple of entries the following examples were found: \mapu{trongongkün} `hit continuously' under the entry for `hit'; \mapu{trongo yüw} `aquiline nose' under the entry for `nose' (maybe `dense nose'); and \mapu{trongongün} `produce hitting sounds, a sound of a hit', \mapu{trongongtrongongngen} `continuous hitting noises' under the entry for `noise', all from de Augusta. Additionally, the entries \mapu{trong} `dense, thick, solid thing' and `frown, make a frown, look with a frown' (maybe `dense forehead') \mapu{lolin, lolikan} were found in Febrés.

Ultimately, \mapu{-künu} and \mapu{-nie} do not contribute to the valency discrimination of verbal roots. Both are applied to verbs of either valency, and the final forms where they are applied are not always transitive, despite both suffixes originating from transitive verbs, as demonstrated by the following example:

\paragraph{\example{34} \label{e34}}~
\vspace{-10pt}
\begin{enumerate}[label=\alph*.]
\item[] 
\mapuex{doy \hspace{24pt} küme \hspace{6pt} monge \hspace{4pt} \textbf{-nie} \hspace{6pt} -tu \hspace{6pt} -a \hspace{10pt} -fu \hspace{6pt} -m} \hfill (\citealp{smeets2008}: 472; 36)\\
\gloss{AV.more AJ.good IV.live \textbf{+PRPS} +RE +FUT +RI +IVN}\\
`have a better life again' -- `tener una mejor vida denuevo'; \textit{literal:} `better living [one] will have again'
\end{enumerate}

\subsection{\label{sec.04.3} Transitivity with \mapu{-(ñ)ma} (Experimentative)}

The final case of suffixes inducing transitivisation in intransitive verbs, reviewed to assess whether these suffixes reveal the valency of root verbs, is \mapu{-(ñ)ma}, the Experimentative suffix as named by \cite{smeets2008}.

\citeauthor{smeets2008} (\citeyear{smeets2008}: 301) states that the morpheme \mapu{-(ñ)ma}, in slot 35 next to the root, denotes a subject experiencing the process of the intransitive verb it accompanies. This suggests that if \mapu{-(ñ)ma} is present, the root is intransitive. However, \citeauthor{zuniga2009b} (\citeyear{zuniga2009b}: 12) argues that \mapu{-(ñ)ma} alternates with the suffix \mapu{-(ü)ñmu} —labelled as `Satisfactive' and placed in slot 25 by \citeauthor{smeets2008} (\citeyear{smeets2008}: 272). The alternation involves aloaffectation, \mapu{-(ñ)ma}, versus autoaffectation, \mapu{-(ü)ñmu}. It remains unclear whether \mapu{-(ñ)ma} is a single morpheme with different functions or if there are distinct morphemes with similar forms. For example, \cite{smeets2008} includes \mapu{-(ü)ñmu} as a satisfactive indicator, \mapu{-(ñ)ma} as an indirect object marker, and another \mapu{-(ñ)ma} as experimentative in different slots, while \mapu{Düngupeyüm} (\citealp{chandia2021}: 46) labels \mapu{-ñma} as an adverbializer, indicating possible overlap.

In the \mapu{longko}'s memoirs \citep{mosbach1930}, forms with the sequence \mapu{ñma} (one variant of the listed suffixes) were examined. A total of 270 words from 108 roots were analysed using \mapu{Düngupeyüm}. The analysis showed that 34 forms have only the indirect object suffix, 43 only the experimental suffix, and 167 can be interpreted with either morpheme, leading to at least two analyses: one with \textsc{IO} and another with \textsc{EXP}. Additionally, one form includes both suffixes, and four contain the adverbial suffix.

Due to the uncertainty surrounding the morpheme \mapu{-(ñ)ma} and the resulting ambiguity in analyses, determining valency based on its occurrence has been avoided. Additionally, the count mentioned earlier used only one allomorph of the suffixes.

Nonetheless, it seems that intransitive roots are rarely transitivised with the experimentative suffix. For instance, in \hyperref[e35a]{E.35a}, transitivisation occurs, while in \hyperref[e35b]{E.35b}, it does not:

\paragraph{\example{35} \label{e35}}~
\vspace{-10pt}
\begin{enumerate}[label=\alph*.]
\item \label{e35a} \mapuex{chew \hspace{24pt} püra \hspace{8pt} \textbf{-ñma} \hspace{2pt} -w \hspace{10pt} -ø \hspace{8pt} -ø \hspace{4pt} -iñ \hspace{6pt} kawell} \hfill \citep{mosbach1930}\\
      \gloss{IP.where IV.climb \textbf{+EXP} +REF +IND +1 +PL NN.horse}\\
      `where do we send your horse?'\\
      \textit{literal:} `where do we make go up your horse?'
\item \label{e35b} \mapuex{fey \hspace{18pt} mew \hspace{4pt} anü \hspace{2pt} \textbf{-ñma} \hspace{2pt} -y \hspace{10pt} -ø \hspace{8pt} ñi \hspace{12pt} witral} \hfill \citep{mosbach1930}\\
      \gloss{DP.that INST IV.sit \textbf{+EXP} +IND +3 SP.her NN.loom}\\
      `there sat in front of her loom'
\end{enumerate}

\section{\label{sec.05} Conclusions}

This study investigates using various suffixes to determine the valency of \mapu{Mapudüngun} verb roots. A reliable computational morphological analyser requires clear, non-redundant data, ideally listing roots under a single classification to minimise ambiguity. However, labile roots are listed under both intransitive and transitive categories, which may introduce some ambiguity but also allows for potential analyses that might otherwise be excluded.

After establishing methods to verify root categories \citep{chandia2025}, we examined suffixes that could indicate verb valency. We assessed the reliability of these suffixes and used the most effective ones to set morphological criteria for recognising verb valency.

We reviewed the following morphemes: aspectual suffixes \mapu{-nie}, \mapu{-künu}, \mapu{-(kü)le} (\S\ref{sec.03.1} and \S\ref{sec.04.1}); adjectiviser \mapu{-fal} (\S\ref{sec.03.2}); causative suffixes \mapu{-(ü)m} (\S\ref{sec.03.3}) and \mapu{-(ü)l} (\S\ref{sec.03.4}); transitiviser/applicative \mapu{-tu} (\S\ref{sec.03.7}); person agreement suffixes \mapu{-fi} (which marks the direct transition) and \mapu{-e} (the inverse marker) (\S\ref{sec.03.5}); and experimental marker \mapu{-(ñ)ma} (\S\ref{sec.04.3}).

The reliability of the tested suffixes can be summarised systematically as follows: \mapu{-(ü)m} and \mapu{-(ü)l} are reliable indicators of intransitivity, as they consistently transitivise intransitive roots. The person agreement suffixes \mapu{-fi} and \mapu{-e} are partially reliable: they indicate transitivity but also appear with labile roots. The following suffixes are unreliable: \mapu{-fal} is too rare and admits an alternative analysis; \mapu{-tu} is polysemous and its valency effects are inconsistent; \mapu{-nie} and \mapu{-künu} apply to roots of either valency class; and \mapu{-(ñ)ma} has ambiguous functions.

Forms ending in \mapu{-fal} involve a transitive root but are rarely used, limiting their effectiveness for identifying root valency (\S\ref{sec.03.2}). The suffix \mapu{-tu} affects valency in various ways, making it unsuitable for determining root valency (\S\ref{sec.03.7}). The person agreement suffixes \mapu{-fi} and \mapu{-e} can facilitate the identification of transitive roots. However, these suffixes also occur with intransitive forms that have been transitivised by other affixes, and with labile verbs, thereby complicating the distinction of valency (\S\ref{sec.03.5}).

The causatives \mapu{-(ü)m} and \mapu{-(ü)l} have been identified as the most effective tools for determining intransitivity; nonetheless, this relies on the confirmation that the roots in question are indeed verbal. Roots transitivised with \mapu{-(ü)m} are listed in Table 4 of \citealp{chandia2025}, while transitivised roots with \mapu{-(ü)l} are illustrated in \citealp{chandia2025}: \S2.3, based on data from the memoirs of the \mapu{longko Paskwal Koña} \citep{mosbach1930}. The use of \mapu{-(ü)m} seems to be decreasing in favour of \mapu{-(ü)l} among current \mapu{Mapudüngun} speakers, though a systematic sociolinguistic study would be required to confirm this impression.

Beyond the main objective, this study explored related topics and proposed alternative analyses, such as reconsidering the morpheme \mapu{-fal} and examining the possibility of labile verbs (\S\ref{sec.03.4.1}). It suggested a diverse origin for homophonous suffixes of the form \mapu{-tu} and proposed that verbal themes occurring with \mapu{küno-} and \mapu{nie-} might be verbal roots forming compounds rather than aspectual suffixes following a main root (\S\ref{sec.04.1}). Further investigation into the habituative suffix \mapu{-ke} (\S\ref{sec.03.6}) is encouraged to uncover potential data on verbal forms.

The morphotactic approach adopted here—inferring root valency from permissible suffix combinations—has clear limitations. First, it cannot resolve cases where a root is genuinely labile and both transitive and intransitive uses are equally frequent without morphological marking. Second, the method depends on the quality and representativeness of the corpus; Koña's memoirs, while extensive, represent a single idiolect from a specific historical period. Third, morphotactics alone cannot distinguish between root valency and valency induced by constructional or discourse factors. Future research should systematically complement morphological analysis with syntactic tests (e.g., passivisation, applicativisation) and semantic criteria. The method proposed here is sufficient for a first-pass classification but requires triangulation with other levels of analysis for problematic cases.

This study is, by necessity, partly exploratory. The complex interaction between root valency, suffixation, and constructional effects means that not all cases could be resolved with certainty. Nevertheless, this article establishes a reliable method for determining intransitive valency using causatives \mapu{-(ü)m} and \mapu{-(ü)l}, which benefits the computational analyser and enhances the understanding of the \mapu{Mapuche} language. Future work will extend the analysis to transitive roots and incorporate syntactic tests to resolve the ambiguous cases identified here.

\newpage
\section*{Acknowledgements}

As was the case in the preceding article, which formed part of the same research work as the present article, I would like to express my gratitude to my thesis supervisors, Irene Castellón and Elisabeth Comelles, for their invaluable guidance in shaping the structure and content of these articles. I would also like to express my sincere appreciation to Fernando Zúñiga, who provided me with invaluable feedback on my claims regarding the \mapu{Mapuche} language, ensuring that they were both comprehensible and firmly rooted in concrete and substantiated facts.

\vspace{2cm}
\bibliographystyle{johd}
\bibliography{bib}

\newpage
\appendix
\section*{Appendix: List of Abbreviations}
\footnotesize
\begin{multicols}{2}
\raggedcolumns
\noindent
1, 2, 3: 1st, 2nd, 3rd person\\
1t2A: 1st to 2nd person Agent\\
3A: 3rd person Agent\\
3P: 3rd person Patient\\
ADJ: Adjectiviser\\
ADJDO: Adjective + Doable\\
AJ: Adjective\\
AV: Adverb\\
BEN: Benefactive\\
CA: Causative\\
CJ: Conjunction\\
COLL: Collective\\
CONT: Continuative\\
DL: Dual\\
DP: Demonstrative Pronoun\\
EXP: Experimentative\\
FAC: Factitive\\
FORCE: Force majeure\\
FUT: Future\\
HAB: Habitual\\
IND: Indicative\\
IND1SG: Indicative 1st person Singular\\
INST: Instrumental\\
INV: Inversion\\
IO: Indirect Object\\
IP: Interrogative Pronoun\\
IV: Intransitive Verb\\
IVN: Intransitive Verbal Noun\\
LOC: Locative\\
MIO: More Involved Object\\
NEG: Negative\\
NN: Noun\\
NOM: Nominaliser\\
NU: Numeral\\
OO: Oblique Object\\
OVN: Objective Verbal Noun\\
PASS: Passive\\
PFPS: Perfective Persistent\\
PL: Plural\\
PLR: Pluraliser\\
PRPS: Progressive Persistent\\
PVN: Plain Verbal Noun\\
RE: Repetitive\\
REF: Reflexive/Reciprocal\\
RI: Ruptured Implicature\\
SFR: Stem Formative\\
SG: Singular\\
SJI: Subjunctive in Imperatives\\
SP: Possessive Pronoun\\
ST: Stative\\
SVN: Subjective Verbal Noun\\
TH: Thither\\
TR: Transitiviser\\
TV: Transitive Verb
\end{multicols}
\end{document}